\pdfoutput=1
\documentclass[11pt]{article}
\usepackage[preprint]{acl}

\usepackage{times}
\usepackage{latexsym}

\usepackage[T1]{fontenc}
\usepackage[utf8]{inputenc}

\usepackage{microtype}
\usepackage{inconsolata}

\usepackage{amsmath}
\usepackage{amsfonts}
\usepackage{amssymb}

\usepackage{graphicx}
\usepackage{subcaption}

\usepackage{booktabs}
\usepackage{multirow}
\usepackage{makecell}

\usepackage{xcolor}
\usepackage{colortbl}
\usepackage{tcolorbox}
\definecolor{Highlight}{rgb}{0.92,0.94,1}

\usepackage{algorithm}
\usepackage[noend]{algpseudocode}

\usepackage{url}
\usepackage{hyperref}

\usepackage{mdframed}

\usepackage{algorithmicx}

\usepackage{tabularx}

\usepackage{color}

\usepackage{CJKutf8}

\usepackage{xspace}

\usepackage{xspace}
\newcommand{\mname}{PlanGPT-VL\xspace}

\title{\mname: Enhancing Urban Planning with Domain-Specific Vision-Language Models}

\author{
He Zhu\textsuperscript{1}\thanks{\ \ Equal contribution.},
Junyou Su\textsuperscript{1}\footnotemark[1],
Minxin Chen\textsuperscript{1}\footnotemark[1],
Wen Wang\textsuperscript{1}, \\
Yijie Deng\textsuperscript{1}, 
Guanhua Chen\textsuperscript{2}, 
Wenjia Zhang\textsuperscript{1}\thanks{\ \ Corresponding author: \texttt{wenjiazhang@pku.edu.cn}} \\
\textsuperscript{1}Behavioral and Spatial AI Lab, Peking University \& Tongji University \\
\textsuperscript{2}Southern University of Science and Technology \\
\texttt{zhuye140@gmail.com,wenjiazhang@pku.edu.cn}
}

\begin{document}
\maketitle

\begin{abstract}
  In the field of urban planning, existing Vision-Language Models (VLMs) frequently fail to effectively analyze and evaluate planning maps, despite the critical importance of these visual elements for urban planners and related educational contexts. Planning maps, which visualize land use, infrastructure layouts, and functional zoning, require specialized understanding of spatial configurations, regulatory requirements, and multi-scale analysis. To address this challenge, we introduce \textbf{\mname}, the first domain-specific Vision-Language Model tailored specifically for urban planning maps. \mname employs three innovative approaches: (1) PlanAnno-V framework for high-quality VQA data synthesis, (2) Critical Point Thinking to reduce hallucinations through structured verification, and (3) comprehensive training methodology combining Supervised Fine-Tuning with frozen vision encoder parameters. Through systematic evaluation on our proposed PlanBench-V benchmark, we demonstrate that \mname significantly outperforms general-purpose state-of-the-art VLMs in specialized planning map interpretation tasks, offering urban planning professionals a reliable tool for map analysis, assessment, and educational applications while maintaining high factual accuracy. Our lightweight 7B parameter model achieves comparable performance to models exceeding 72B parameters, demonstrating efficient domain specialization without sacrificing performance.
\end{abstract}

\section{Introduction}

Vision-Language Models (VLMs) have demonstrated remarkable progress in general multimodal tasks, including image understanding \cite{gpt4o,gemini}, visual reasoning \cite{zhu2025internvl3,guo2025seed1}, and multimodal dialogue \cite{liu2023llava, wang2024qwen2vl}. Recent research has successfully extended these models to specialized domains such as medical imaging \cite{li2023llava, lai2025med, pan2025medvlm}, geographical information systems \cite{zhang2024bb, zhang2024mapgpt}, and mathematical reasoning \cite{chen2025bring, shen2025vlm}, with corresponding domain-specific benchmarks emerging in aesthetics \cite{huang2024aesbench, zhou2024uniaa, lin2024designprobe}, autonomous driving \cite{qian2024nuscenes, sima2024drivelm}, and other fields. Despite these advances, we identify urban planning as a critical domain that could significantly benefit from specialized VLMs to \textit{interpret complex planning maps}—a task where even leading commercial models exhibit substantial limitations in recognizing specialized elements and applying the cartographic interpretation skills essential for planning practices.

\begin{figure}[t]
  \centering
  \includegraphics[width=0.5\textwidth]{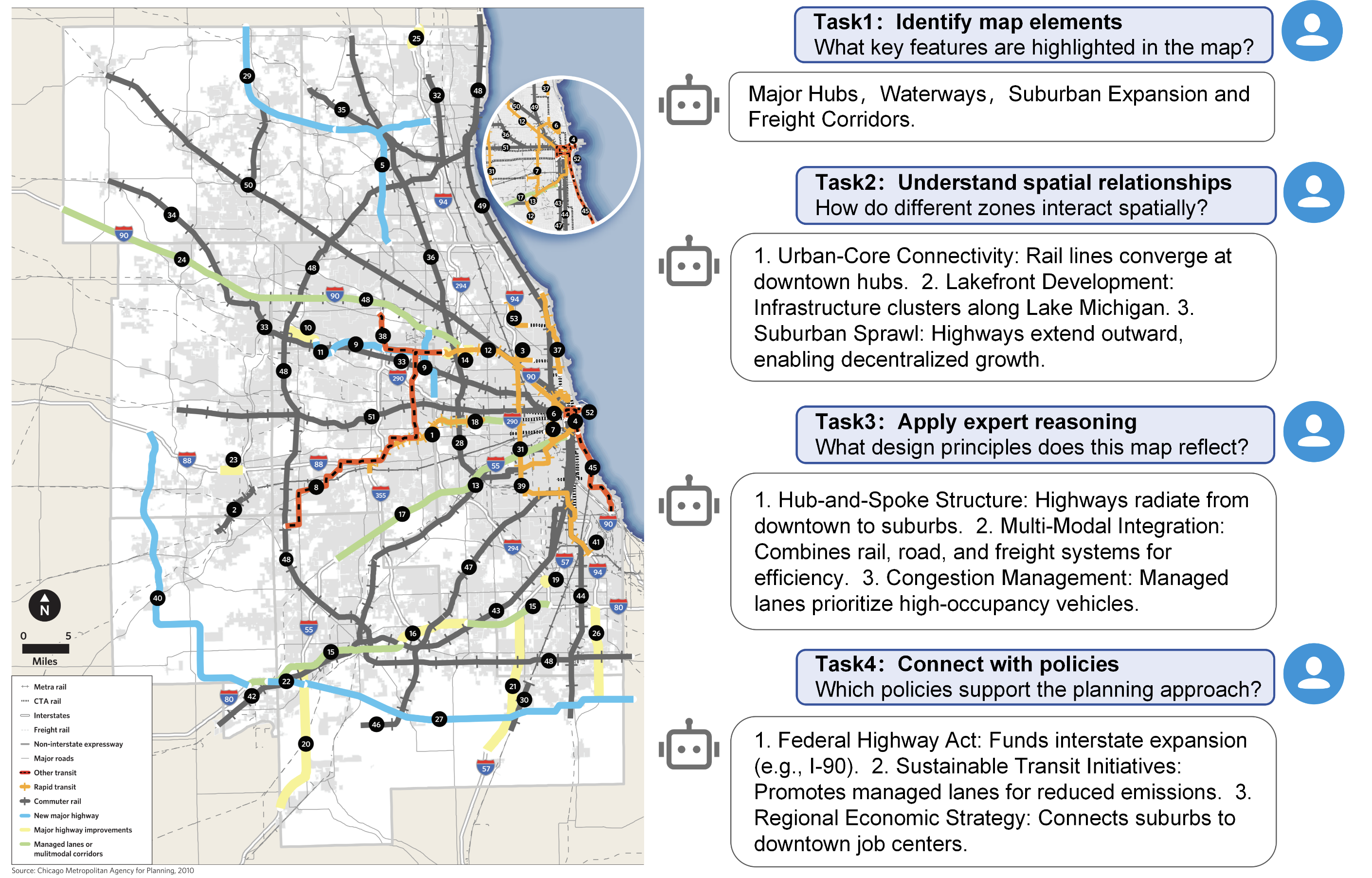}
  \caption{Urban planning multimodal tasks including map elements identification, spatial relationships understanding, expert reasoning, policy association and other key applications.}
  \label{fig:usage1}
\end{figure}

Planning maps are essential tools in urban development that visually represent current conditions, future plans, and policy guidelines. Unlike general maps, planning maps employ specialized symbols, color-coding systems, and annotations to indicate land use zones, transportation networks, and development restrictions \cite{Lynch1984, Steinitz1995, Healey1997}. As illustrated in Figure~\ref{fig:usage1}, urban planning involves multiple types of multimodal tasks including map elements identification, spatial relationships understanding, expert reasoning, policy association, etc. Modeling human mobility patterns is essential for urban planning and policy evaluation. Current general-purpose VLMs face three critical limitations when applied to these tasks: (1) High hallucination rates in information-dense planning maps, where models frequently misidentify zones and fabricate non-existent features; (2) Responses that don't align with urban planners' preferred professional language and communication styles; and (3) Unreliable evaluation methods for objectively assessing specialized planning map interpretation. These challenges largely stem from the scarcity of domain-specific visual question-answering data and the prohibitive cost of manual annotation by planning experts \cite{liu2024mminstruct,llava-med}.

In this paper, we introduce \textbf{\mname}, the first domain-specific Vision-Language Model tailored specifically for urban planning map interpretation. Our approach addresses the identified challenges through a comprehensive technical framework centered on three key innovations. First, we develop the \textbf{PlanAnno-V framework} for synthesizing high-quality instruction-response pairs through domain-specific data preprocessing with expert annotation, systematic instruction synthesis that preserves professional expertise while expanding distributional coverage, and model-specific rewriting to align with professional communication patterns. Second, we introduce \textbf{Critical Point Thinking (CPT)}, a novel methodology that reduces hallucinations by decomposing complex visual planning information into verifiable critical points and employing a 'Generate-Verify-Revise' paradigm. Third, we create \textbf{PlanBench-V}, the first comprehensive benchmark for evaluating VLM performance on urban planning map interpretation tasks, professionally annotated by urban planning experts. Additionally, we explore the trade-off between general capabilities and domain-specific expertise, providing insights into optimal model specialization strategies for urban planning applications. Our experiments demonstrate that \mname outperforms both open-source and commercial VLMs by an average of 59.2\% on specialized planning map interpretation tasks, with our lightweight 7B parameter model achieving comparable performance to models exceeding 72B parameters. Our contributions include: (1) Introduction of \mname-2B/7B, the first specialized VLM for urban planning that achieves advance performance while maintaining a compact model size; (2) Development of the PlanAnno-V framework that efficiently generates high-quality training data; and (3) Creation of PlanBench-V for systematic evaluation of planning map interpretation capabilities.

\section{Related Works}

\paragraph{Domain-Specific Language and Vision-Language Models}
Large language models have evolved from general-purpose systems \cite{GPT4,chatgpt,touvron2023llama,PaLM2,Claude2,mistral,gemini} to specialized applications across diverse domains. In the Chinese language context, models such as DeepSeek \cite{deepseekai2025deepseekv3technicalreport}, Baichuan \cite{baichuan2}, GLM \cite{GLM}, and Qwen \cite{qwen2025qwen25technicalreport} have addressed specific linguistic requirements. Domain adaptation has produced specialized systems in medicine (HuaTuo \cite{huatuo}, DoctorGLM \cite{Doctorglm}), legal (ChatLaw \cite{ChatLaw}), finance (XuanYuan 2.0 \cite{XuanYuan}), and mathematics (MathGPT \cite{MathGPT}). Several models address aspects of urban environments, including PlanGPT \cite{zhu2024plangpt} for text-based urban planning, TrafficGPT \cite{TrafficGPT} for transportation management, NASA's Prithvi \cite{Prithvi-100M} for climate predictions, and CityGPT \cite{feng2024citygpt} for spatial reasoning. However, none specifically addresses the visual interpretation of planning maps with their specialized representational requirements and domain-specific reasoning needs, which motivates our development of \mname as the first vision-language model specifically designed for urban planning map interpretation.

\paragraph{Multimodal Instruction Data Synthesis}
VLM effectiveness relies on high-quality instruction-response pairs, with recent work developing synthetic data pipelines similar to our PlanAnno-V framework. General-purpose approaches like MAmmoTH-VL \cite{guo2024mammoth}, MMInstruction \cite{liu2024mm}, and Infinity-Multimodal \cite{gu2024infinity} employ template-based sampling and clustering techniques, while OASIS \cite{zhang2025oasis} uses visual prompting for grounded instructions. Most relevant to our Critical Point Thinking methodology, MM-Verify \cite{sun2025mm} introduces verification mechanisms, and LLaVA-CoT \cite{xu2025llavacot} implements structured reasoning synthesis. Our PlanAnno-V framework extends these approaches through domain-specific preprocessing, professional language alignment, and specialized verification suited to the complex visual reasoning demands of urban planning maps.

\begin{figure*}[t]
  \centering
  \includegraphics[width=0.95\textwidth]{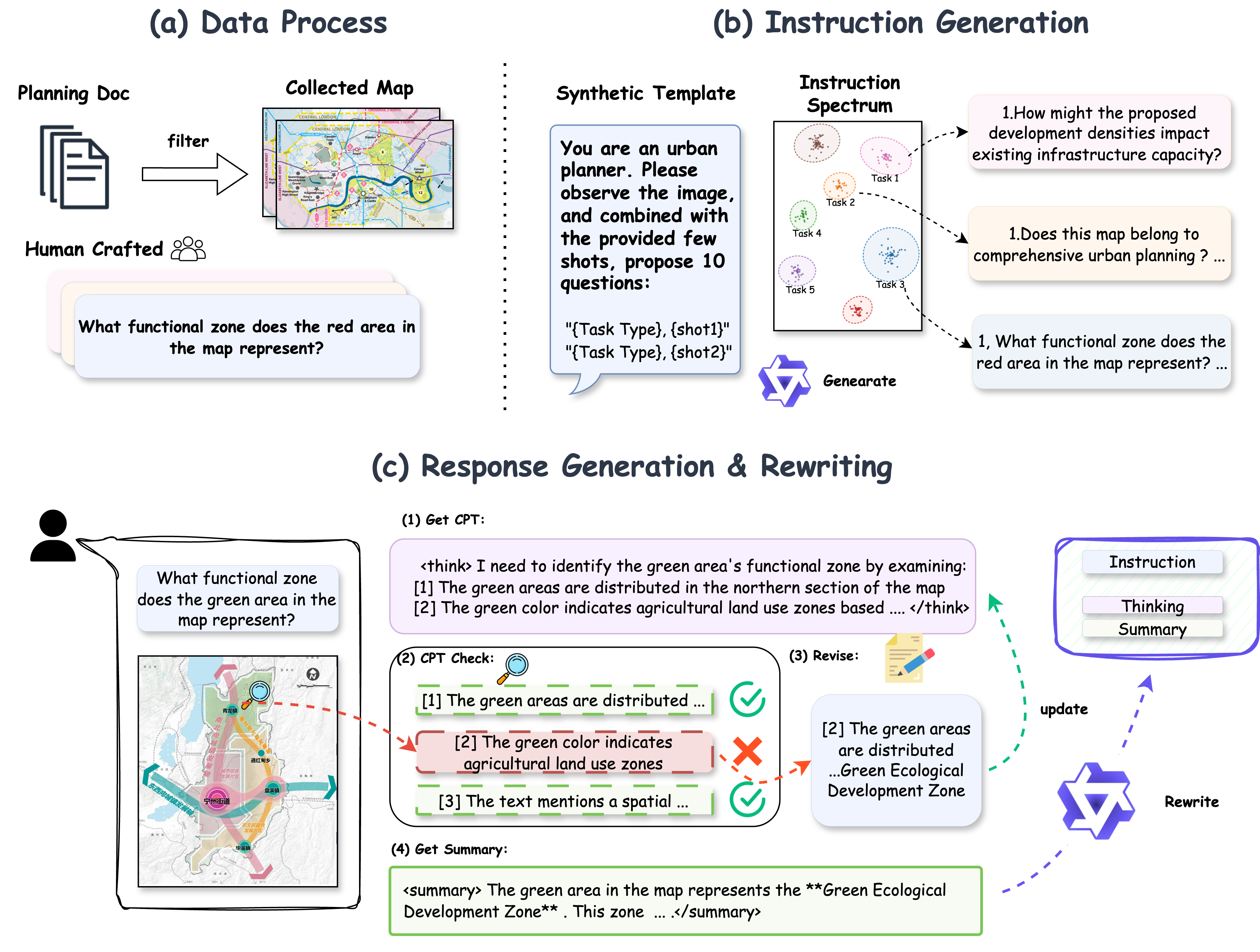}
  \caption{Overview of PlanAnno-V framework. Our approach synthesizes high-quality instruction-response pairs through a three-stage process: (1) domain-specific data preprocessing with expert annotation, (2) instruction-response synthesis using Critical Point Thinking for hallucination reduction, and (3) model-specific rewriting to align with professional planning communication patterns.}
  \label{fig:plananno}
\end{figure*}

\section{Method} \label{sec:framework}

To address the challenges of specialized planning map interpretation and the scarcity of high-quality training data, we introduce the PlanAnno-V framework. This comprehensive approach centers on three key innovations: (1) A systematic data synthesis pipeline for generating high-quality instruction-response pairs, as shown in section \ref{sec:overview}; (2) Critical Point Thinking (CPT) for hallucination mitigation, as shown in section \ref{sec:cpt}; and (3) PlanBench-V for reliable evaluation of planning-specific capabilities, as shown in section \ref{sec:planbench}.

\subsection{Overview of PlanAnno-V} \label{sec:overview}

The PlanAnno-V framework aims to synthesize high-quality, low-hallucination visual instruction tuning data for enhancing domain-specific model capabilities with minimal human intervention. As depicted in Figure~\ref{fig:plananno}, PlanAnno-V takes unlabeled documents containing planning maps as input and outputs professional-quality visual instruction-response pairs through three stages: \textbf{Stage 1: Domain-Specific Data Preprocessing} involves collecting and filtering planning maps followed by expert annotation of seed data. We collected approximately 5,000 maps from urban planning bureaus, then applied diversity-based filtering to select 1,050 representative maps with maximum visual and content variation. Domain experts manually annotated around 800 high-quality examples from a subset of approximately 50 maps selected from this filtered set, providing professional seed data for subsequent stages. Further details are presented in Appendix~\ref{appendix:data_preprocessing}. \textbf{Stage 2: Instruction-Response Synthesis} combines diversity-enhanced instruction generation (Section \ref{sec:instruction_synthesis}) that preserves professional expertise while systematically expanding distributional coverage, and Critical Point Thinking (CPT) (Section \ref{sec:cpt}) for verifiable response synthesis that reduces hallucination through structured verification. \textbf{Stage 3: Model-Specific Rewriting} employs target models to align responses with professional planning communication styles, incorporating planner examples as in-context demonstrations to ensure domain-appropriate linguistic patterns.

\subsection{Distributional Instruction Synthesis} \label{sec:instruction_synthesis}

Manual annotation by domain experts yields high-quality, professionally relevant instructions but inherently suffers from limited diversity and complexity coverage \cite{wang2022self}. Our approach preserves this professional expertise while systematically expanding the distributional coverage through principled synthesis methods.

\paragraph{Instruction Spectrum Construction}
We begin with 1k professionally curated instructions from urban planning experts. Each instruction undergoes automated intent extraction via InstaTagger \cite{lu2023instag}, which identifies semantic components underlying the query. For example, "Identify ecological protection red lines and analyze their impact on residential development" yields intents "spatial\_analysis" and "location\_identification". Through clustering analysis, we categorize instructions into 8 distinct task types and establish a complexity hierarchy based on average intent count per type, creating a comprehensive instruction spectrum across the urban planning domain.

\paragraph{Systematic Distributional Expansion}
Inspired by recent advances in automated instruction generation \cite{liu2024mminstruct, luo2024mmevolempoweringmultimodallarge, zhu2024fanno}, we implement a stratified replication strategy to expand beyond seed data limitations. For each planning image $i$, we sample diverse task types $\mathcal{T} = \{t_1, t_2, ..., t_{10}\}$ and their corresponding exemplars $\mathcal{E} = \{e_1, e_2, ..., e_{10}\}$ as few-shot demonstrations. The instruction generation process is formalized as: $p(q_{new} | i, \mathcal{T}, \mathcal{E}) = \text{Generate}(i | \{(t_j, e_j)\}_{j=1}^{10}, \phi_{div})$ where $q_{new}$ represents the synthesized instruction, and $\phi_{div}$ denotes diversification prompts that encourage task variety and complexity progression. 

\subsection{Critical Point Thinking for Hallucination Mitigation} \label{sec:cpt}

Hallucination presents a fundamental challenge when interpreting complex visual content. Inspired from \citet{yu2024rlaif}, we introduce \textit{Critical Point Thinking} (CPT), which decomposes reasoning into structured, verifiable components to systematically reduce factual errors through iterative verification and correction.

\begin{algorithm}[h]
\caption{Critical Point Thinking (CPT)}
\label{alg:cpt}
\begin{algorithmic}[1]
\Require Planning map $m$, instruction $i$, verification threshold $\tau$
\Ensure Verified response $r_{final}$
\State $\mathcal{P} \leftarrow \text{ExtractCriticalPoints}(m, i)$ 
\For{each $p_j \in \mathcal{P}$}
   \State $q_j \leftarrow \text{FormulateVerificationQuery}(p_j, i)$
   \State $v_j \leftarrow \text{VerifyPoint}(q_j, m)$ 
   \If{$v_j < \tau$}
       \State $p_j \leftarrow \text{CorrectPoint}(p_j, m, q_j)$ 
   \EndIf
\EndFor
\State $\mathcal{P}_{merged} \leftarrow \text{MergeRedundantPoints}(\mathcal{P})$
\State $r_{final} \leftarrow \text{ReconstructResponse}(\mathcal{P}_{merged})$
\State \Return $r_{final}$
\end{algorithmic}
\end{algorithm}

As depicted in Table~\ref{alg:cpt}, our CPT framework employs a systematic ``Generate-Verify-Revise`` paradigm. The key insight is that models excel at focused verification tasks compared to open-ended generation, especially when attention is concentrated on specific visual elements \cite{gpt4o}. We first extract critical points in structured format ("Critical Point 1: ...", "Critical Point 2: ..."), then verify each atomic claim through targeted queries against the planning map and correct any identified errors. Finally, we review the critical points to eliminate redundancy, mitigating the problem of overthinking \cite{chen2025think23overthinkingo1like}. Subsequent ablation experiments in Section~\ref{tab:ablation_exp} demonstrate the effectiveness of our approach.

\subsection{PlanBench-V: A Benchmark for Urban Planning VLMs} \label{sec:planbench}

To address the challenge of evaluating domain-specific visual understanding in urban planning, we introduce PlanBench-V, the first comprehensive benchmark for assessing VLM performance on planning map interpretation tasks. PlanBench-V consists of 300 carefully curated examples spanning diverse planning tasks including zoning analysis, infrastructure assessment, spatial reasoning, and regulatory compliance, with categorical distribution illustrated in Figure \ref{fig:instruction_analysis}(c). Each example is annotated by three professional urban planners with specific evaluation criteria.

To overcome the inherently open-ended nature of planning inquiries, we establish a multi-dimensional scoring framework where each question is associated with $n$ expert-defined evaluation criteria $\{c_1, c_2, ..., c_n\}$. For automated assessment, we employ a specialized evaluation protocol that computes a normalized score $S = \frac{\sum_{i=1}^{n} \mathbb{I}(c_i \in R)}{n}$, where $\mathbb{I}(\cdot)$ is the indicator function denoting criteria satisfaction. This approach enables objective evaluation of subjective planning insights while maintaining alignment with professional standards. The comprehensive evaluation protocol is documented in Appendix \ref{appendix:planvlm_eval} and evaluation eamples in Appendix\ref{appendix:planvlm_eval_example}.

\subsection{Training Methodology} \label{sec:training_methodology}
Our training process employs a streamlined approach that balances efficiency and effectiveness. We implemented \mname by fine-tuning the Qwen2-7B-Instruct model \cite{wang2024qwen2vl} while freezing both the vision encoder and projector layers to preserve general visual understanding capabilities. This freezing strategy was adopted after our experiments revealed that fine-tuning these components led to more severe degradation of general capabilities, as detailed in Section~\ref{sec:domain_general_analysis}. We conduct Supervised Fine-Tuning using single-image QA pairs and multi-turn dialogues generated through our PlanAnno-V framework. At inference time, we employ rejection sampling to enhance response quality and stability.

\section{Experiments}
\label{sec:experiments}

\subsection{Experimental Setup}

\paragraph{Implementation Details} We conducted experiments using 4 NVIDIA A100 GPUs (80GB each). Our implementation is based on the VERL framework \cite{sheng2024hybridflow} for efficient VLM fine-tuning. We employed Qwen2-7B-VL-Instruct \cite{wang2024qwen2} as our base model and conducted supervised fine-tuning without a pre-training phase \cite{karamcheti2024prismaticvlmsinvestigatingdesign}. For SFT, we used the AdamW optimizer with a learning rate of 2e-5, cosine learning rate scheduler with 5\% warmup steps, and trained for 3 epochs with a global batch size of 128 and maximum sequence length of 8192 tokens.

\paragraph{Datasets} Our training corpus consists of approximately 10k instruction-following examples generated from 1k selected urban planning maps using our PlanAnno-V framework. The dataset spans multiple query categories including zoning analysis, infrastructure assessment, spatial reasoning, policy compliance, and regulatory alignment. The maps primarily originate from diverse Chinese cities, ensuring geographical coverage. Detailed data analysis and validation procedures are documented in Appendix~\ref{appendix:data_analysis}.

\paragraph{Evaluation Benchmarks} We evaluate our model using both domain-specific and general benchmarks: (1) \textbf{PlanBench-V}, our newly created benchmark described in Section~\ref{sec:planbench}; and (2) \textbf{General VLM Benchmarks} including MMMU \cite{yue2024mmmu}, GQA \cite{hudson2019gqa}, and POPE \cite{li2023evaluating} to assess preservation of general visual understanding capabilities. We use the lmms-eval framework \cite{lmms_eval2024} for standardized evaluation.

\begin{table*}[t]
 \centering
 \resizebox{\textwidth}{!}{
 \begin{tabular}{l|cccccccc|cccc|c}
 \toprule
 \multirow{2}{*}{\textbf{Model}} & \multicolumn{8}{c|}{\textbf{PlanBench-V (Detailed Categories)}} & \multicolumn{4}{c|}{\textbf{PlanBench-V (Main Categories)}} & \multirow{2}{*}{\textbf{Overall}} \\
  & \textbf{Element} & \textbf{Eval} & \textbf{Class} & \textbf{Assoc} & \textbf{Spatial} & \textbf{Prof} & \textbf{Desc} & \textbf{Dec} & \textbf{Perc} & \textbf{Reas} & \textbf{Assoc} & \textbf{Impl} & \\
 \midrule
 \multicolumn{14}{c}{\textit{General Vision-Language Models}} \\
 \midrule
 Qwen2-VL-2B-Instruct & 0.744 & 0.537 & 0.948 & 0.926 & 0.500 & 0.656 & 0.925 & 0.792 & 0.767 & 0.664 & 0.926 & 0.616 & 0.731 {\color{red}(-0.179)} \\
 Qwen2-VL-7B-Instruct & 0.902 & 0.857 & 1.031 & 0.979 & 0.716 & 0.943 & 1.386 & 0.657 & 0.964 & 0.878 & 0.979 & 0.795 & 0.910
 {\color{blue}(base)}\\
 Qwen2-VL-72B-Instruct-AWQ & 1.010 & 0.670 & 1.125 & 1.114 & 0.746 & 0.967 & 1.367 & 0.632 & 1.056 & 0.920 & 1.114 & 0.658 & 0.963 {\color{olive}(+0.053)} \\
 Qwen2.5-VL-3B-Instruct & 0.862 & 0.697 & 0.953 & 0.970 & 0.691 & 0.870 & 1.554 & 0.936 & 0.951 & 0.822 & 0.970 & 0.771 & 0.876 {\color{red}(-0.034)} \\
 Qwen2.5-VL-7B-Instruct & 1.101 & 0.802 & 1.089 & 1.069 & 0.865 & 1.110 & 1.628 & 1.054 & 1.168 & 1.013 & 1.069 & 0.880 & 1.050 {\color{olive}(+0.140)} \\
Qwen2.5-VL-32B-Instruct & 1.432 & 1.678 & 1.578 & 1.685 & 1.539 & 1.791 & 1.928 & 1.620 & 1.496 & 1.649 & 1.685 & 1.660 & 1.616 {\color{blue}(teacher)}\\
 Qwen2.5-VL-72B-Instruct-AWQ & 1.299 & 1.153 & 1.406 & 1.253 & 1.248 & 1.263 & 1.825 & 1.090 & 1.366 & 1.289 & 1.253 & 1.134 & 1.288 {\color{olive}(+0.378)} \\
 \midrule
 InternVL3-8B & 0.992 & 0.631 & 1.026 & 0.926 & 0.798 & 0.751 & 1.783 & 1.073 & 1.094 & 0.831 & 0.926 & 0.768 & 0.909 {\color{red}(-0.001)} \\
 InternVL3-9B & 1.173 & 0.921 & 1.297 & 1.297 & 1.260 & 1.435 & 1.878 & 0.903 & 1.263 & 1.339 & 1.297 & 0.916 & 1.271 {\color{olive}(+0.361)} \\
 InternVL3-14B & 0.931 & 0.709 & 1.177 & 1.098 & 0.793 & 0.998 & 1.580 & 0.917 & 1.014 & 0.962 & 1.098 & 0.773 & 0.980 {\color{olive}(+0.070)} \\
 \midrule
 GPT-4o-mini & 0.664 & 0.636 & 1.021 & 0.963 & 0.789 & 1.030 & 0.890 & 1.175 & 0.693 & 0.938 & 0.963 & 0.803 & 0.866 {\color{red}(-0.044)} \\
 GPT-4o & 1.051 & 1.223 & 1.260 & 1.527 & 1.305 & 1.564 & 1.708 & 1.429 & 1.136 & 1.399 & 1.527 & 1.287 & 1.342 {\color{olive}(+0.432)} \\
 \midrule
 \multicolumn{14}{c}{\textit{Our Models}} \\
 \midrule
 \rowcolor{blue!10} \mname-2B & 1.174 & 1.305 & 1.219 & 1.453 & 1.328 & 1.485 & 1.744 & \textbf{1.567} & 1.247 & 1.366 & 1.453 & 1.386 & 1.352 {\color{olive}(+0.442)} \\
 \rowcolor{blue!10} \mname-7B & \textbf{1.417} & \textbf{1.528} & \textbf{1.541} & \textbf{1.537} & \textbf{1.569} & \textbf{1.729} & \textbf{2.000} & 1.501 & \textbf{1.492} & \textbf{1.627} & \textbf{1.537} & \textbf{1.520} & \textbf{1.566} {\color{olive}(+0.656)} \\
 \bottomrule
 \end{tabular}
 }
 \caption{Performance comparison on PlanBench-V with detailed and main categories. Detailed categories: Element = Element Recognition, Eval = Evaluation, Class = Classification, Assoc = Association, Spatial = Spatial Relations, Prof = Professional Reasoning, Desc = Description, Dec = Decision Making. Main categories: Perc = Perception, Reas = Reasoning, Assoc = Association, Impl = Implementation.}
 \label{tab:planbench_detailed}
\end{table*}

\subsection{Main Results}

We evaluate \mname against state-of-the-art VLMs on both domain-specific planning tasks. Table \ref{tab:planbench_detailed} presents these comprehensive results.

The PlanBench-V evaluation reveals several key patterns: (1) Among general-purpose models, the Qwen series demonstrates superior performance, with Qwen2.5-VL-32B-Instruct achieving the highest score (1.616), likely due to its stronger multimodal alignment and Chinese language capabilities relevant to our dataset. AWQ quantized versions show substantial degradation, with the 72B-AWQ model (1.288) underperforming the unquantized 32B variant, suggesting that aggressive quantization compromises fine-grained visual reasoning needed for planning tasks. (2) Proprietary models like GPT-4o (1.342) underperform compared to leading open-source alternatives, potentially reflecting training biases toward general rather than domain-specific visual content. (3) Across all models, performance is generally stronger in Description and Classification tasks but weaker in Evaluation and Decision Making dimensions, which require deeper domain expertise. (4) \mname-7B achieves the top overall performance (1.566), demonstrating notable advantages in Professional Reasoning (1.729) and Implementation (1.520) dimensions where domain-specific training provides significant benefit, while maintaining competitive performance across all task categories.

\subsection{Ablation Studies}
We conduct extensive ablation studies to understand the contribution of each component in the PlanAnno-V framework. Table~\ref{tab:ablation_exp} presents a comprehensive comparison of different model variants, evaluating both PlanBench-V performance (domain-specific) and general vision-language benchmarks. All PlanBench-V scores are normalized with GPT-4V as the reference baseline (1.0).

\begin{table*}[t]
  \centering
  \resizebox{\textwidth}{!}{
  \begin{tabular}{lccccc|ccc}
  \toprule
  \multirow{2}{*}{\textbf{Model Variant}} & \multicolumn{5}{c|}{\textbf{PlanBench-V (Domain-Specific)}} & \multicolumn{3}{c}{\textbf{General Benchmarks}} \\
  \cmidrule(lr){2-6} \cmidrule(lr){7-9}
  & \textbf{Perception} & \textbf{Reasoning} & \textbf{Association} & \textbf{Implementation} & \textbf{Overall} & \textbf{MMMU} & \textbf{GQA} & \textbf{POPE} \\
  \midrule
  \textbf{Baseline Model} & & & & & & & & \\
  \midrule
  Qwen2-VL-7B-Instruct & 0.964 & 0.878 & 0.979 & 0.795 & 0.910 & \textbf{51.6} & \textbf{62.3} & 88.3 \\
  \midrule
  \textbf{Qwen2.5-72B-VL-Instruct Teacher Models} & & & & & & & & \\
  \midrule
  + CoT & 1.172 {\color{olive}(+0.208)} & 1.159 {\color{olive}(+0.281)} & 1.108 {\color{olive}(+0.129)} & 0.904 {\color{olive}(+0.109)} & 1.129 {\color{olive}(+0.219)} & 48.3 {\color{red}(-3.3)} & 61.5 {\color{red}(-0.8)} & 88.7 {\color{olive}(+0.4)} \\
  + CPT & 1.231 {\color{olive}(+0.267)} & 1.196 {\color{olive}(+0.318)} & 1.013 {\color{olive}(+0.034)} & 0.990 {\color{olive}(+0.195)} & 1.155 {\color{olive}(+0.245)} & 49.4 {\color{red}(-2.2)} & \textbf{62.3} {\color{olive}(+0.0)} & \textbf{89.0} {\color{olive}(+0.7)} \\
  + CPT + Verification & 1.326 {\color{olive}(+0.362)} & 1.225 {\color{olive}(+0.347)} & 1.172 {\color{olive}(+0.193)} & 1.180 {\color{olive}(+0.385)} & 1.238 {\color{olive}(+0.328)} & 49.1 {\color{red}(-2.5)} & 61.5 {\color{red}(-0.8)} & 88.7 {\color{olive}(+0.4)} \\
  \midrule
  \textbf{Qwen2.5-32B-VL-Instruct Teacher Models} & & & & & & & & \\
  \midrule
  + CPT & 1.464 {\color{olive}(+0.500)} & 1.580 {\color{olive}(+0.702)} & \textbf{1.628} {\color{olive}(+0.649)} & 1.464 {\color{olive}(+0.669)} & 1.547 {\color{olive}(+0.637)} & 46.3 {\color{red}(-5.3)} & 60.5 {\color{red}(-1.8)} & 88.5 {\color{olive}(+0.2)} \\
  \rowcolor{blue!10} + CPT + Verification & \textbf{1.492} {\color{olive}(+0.528)} & \textbf{1.627} {\color{olive}(+0.749)} & 1.537 {\color{olive}(+0.558)} & \textbf{1.520} {\color{olive}(+0.725)} & \textbf{1.566} {\color{olive}(+0.656)} & 45.8 {\color{red}(-5.8)} & 56.0 {\color{red}(-6.3)} & 88.5 {\color{olive}(+0.2)} \\
  \bottomrule
  \end{tabular}
  }
  \caption{Comprehensive ablation study results on PlanBench-V and general vision-language benchmarks.}
  \label{tab:ablation_exp}
\end{table*}

Our ablation analysis reveals several key insights about the PlanAnno-V framework components. First, replacing standard Chain-of-Thought (CoT) with Critical Point Thinking (CPT) yields modest improvements (+2.3\% with 72B teacher), but CPT's main advantage lies in its compatibility with verification mechanisms. Adding the verification component significantly improves performance (+7.2\% overall), particularly benefiting the Implementation dimension (+19.2\%), which validates our hypothesis that iterative verification effectively reduces hallucinations. The most substantial gain comes from upgrading the teacher model quality—moving from 72B to 32B teacher models provides a +25.0\% overall improvement, with particularly strong gains in Reasoning (+29.0\%) and Association (+38.9\%) tasks. However, domain-specific fine-tuning creates a trade-off with general capabilities. MMMU scores drop by 11.2\% (51.6 → 45.8) and GQA by 10.1\%, while POPE remains stable (88.3 → 88.5), suggesting preserved object detection. Our method achieves a significant +72.1\% improvement on domain tasks, but at the cost of general capabilities. We aim to address this limitation through mixed training strategies incorporating general benchmark data.
\section{Analysis}

\subsection{Data Leakage Analysis}

To ensure evaluation validity, we conducted data leakage analysis between our training corpus and evaluation datasets, addressing concerns about contamination in large-scale model development \cite{du2023mods}. We focused on detecting potential image duplicates by extracting CLIP-ViT-L/32 \cite{openai2021clip} embeddings from all images and computing cosine similarities between pairs. Images exceeding a 0.9 similarity threshold underwent manual inspection. As shown in Table \ref{tab:leakage}, we identified minimal high-similarity pairs, and human verification confirmed these were distinct content rather than duplicates. This analysis confirms our performance improvements represent genuine domain-specific capabilities rather than memorization artifacts.

\begin{table}[h]
\centering
\resizebox{0.9\columnwidth}{!}{
\begin{tabular}{lccc}
\toprule
\textbf{Dataset} & \textbf{Total Images} & \textbf{High Similarity} & \textbf{Verified Leakage} \\
\midrule
Seed Data & 50 & 0 (0.00\%) & 0 (0.00\%) \\
PlanAnno-V & 1k & 9 (0.09\%) & 0 (0.00\%) \\
\bottomrule
\end{tabular}
}
\caption{Data leakage analysis showing minimal image duplication.}
\label{tab:leakage}
\end{table}

\subsection{Instruction Synthesis Effectiveness Analysis} 
To validate the quality and diversity of our automated instruction generation pipeline, we conducted a comprehensive comparative analysis between human-annotated seed data and PlanAnno-V synthesized instructions across three critical dimensions: distributional alignment, categorical diversity, and complexity preservation.

\begin{figure*}[t]
  \centering
  \begin{tabular}{ccc}
    \includegraphics[width=0.3\textwidth]{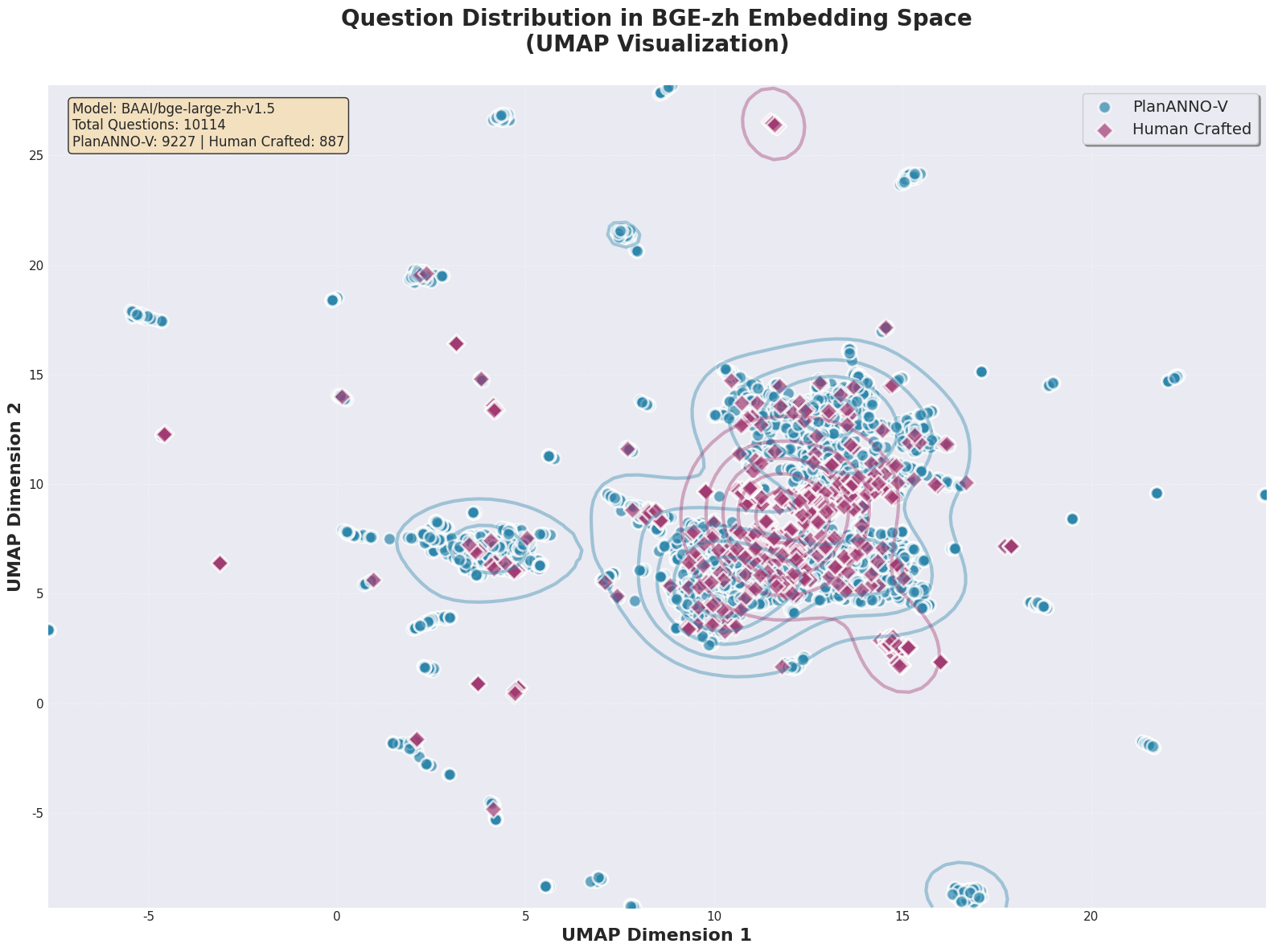} &
    \includegraphics[width=0.3\textwidth]{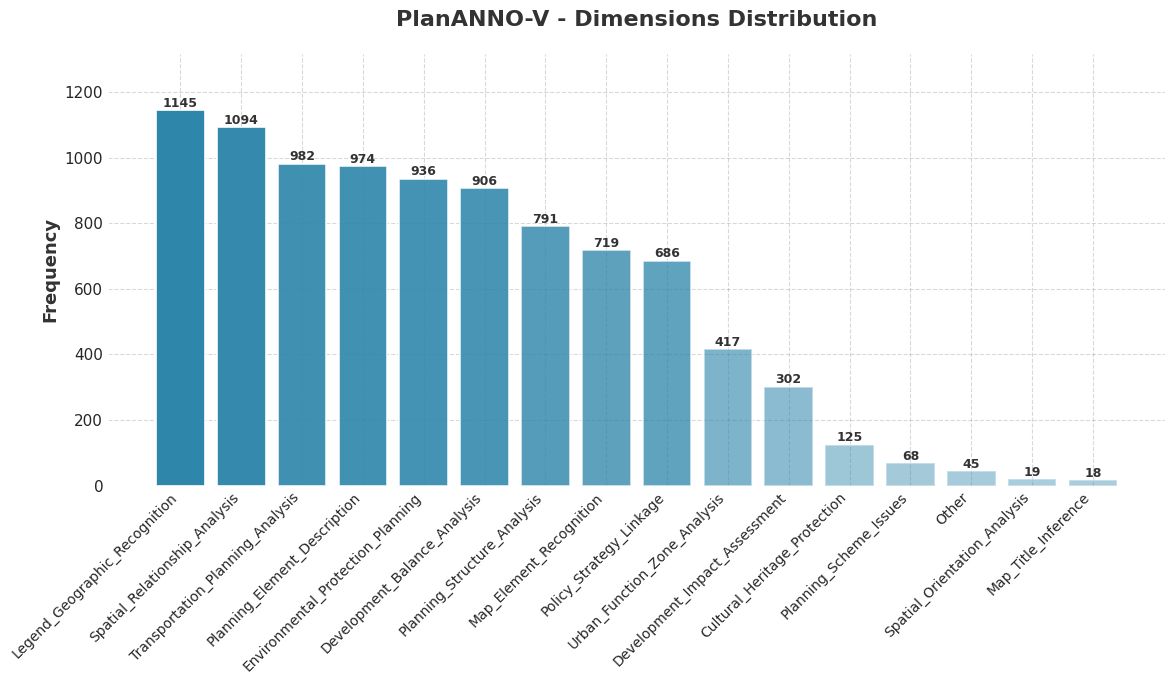} &
    \includegraphics[width=0.3\textwidth]{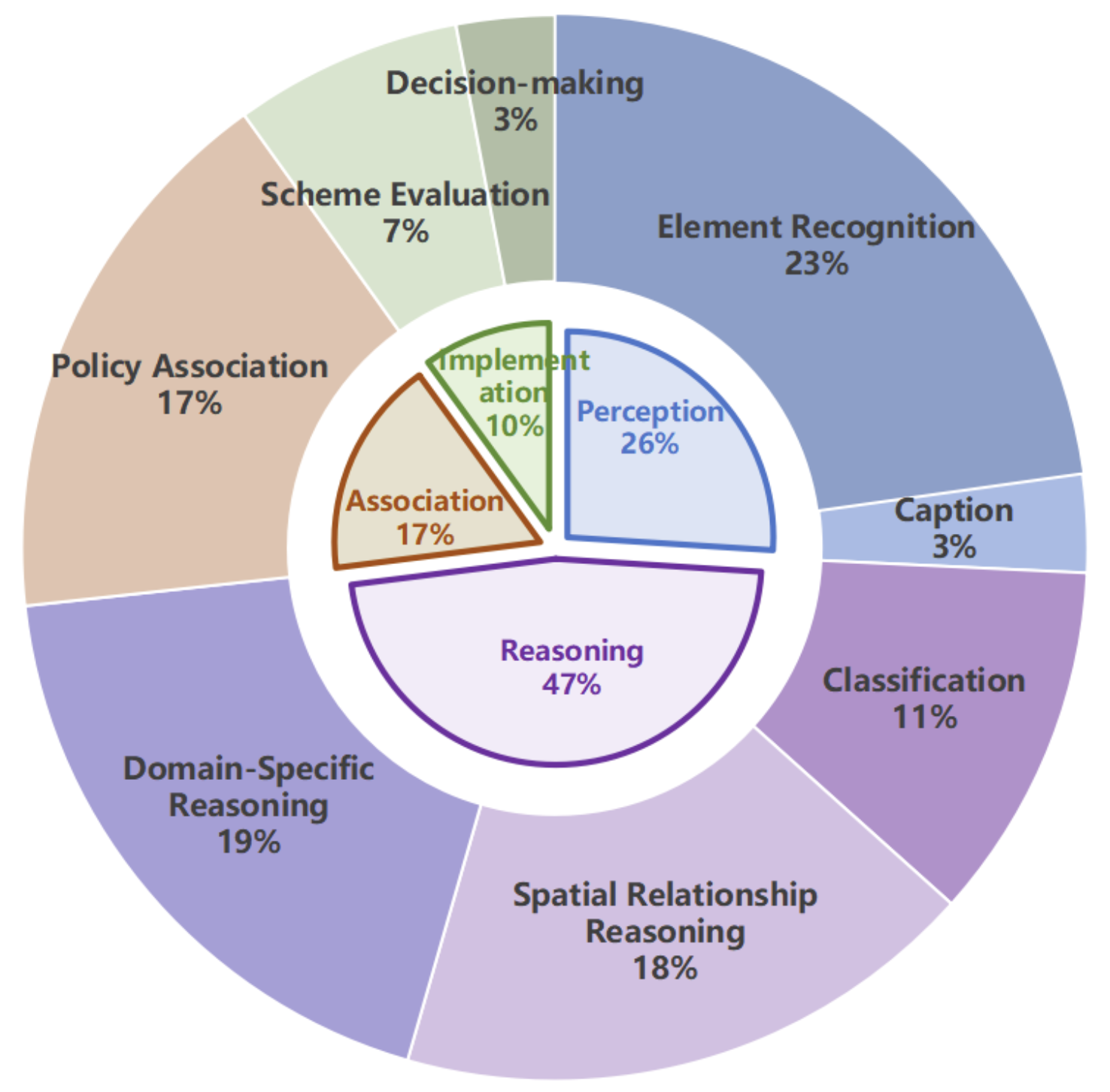} \\
    (a) & (b) & (c)
  \end{tabular}
  \caption{Analysis of PlanAnno-V instruction synthesis: (a) UMAP projection of instruction embeddings with kernel density estimation contours, showing how synthesized instructions (blue) maintain similar distribution patterns to expert-annotated seed data (red) while introducing beneficial diversity; (b) Categorical distribution of synthesized instructions across planning dimensions; (c) Statistical distribution of PlanBench-V Dataset.}
  \label{fig:instruction_analysis}
\end{figure*}

\paragraph{Distributional Alignment} We analyze whether synthesized instructions maintain the underlying distributional characteristics of expert-curated examples. Using the bge-zh-base model \cite{xiao2024c} to map instructions to embeddings, we compute distributional metrics between seed and synthesized data. Our analysis reveals strong semantic alignment (Cosine Similarity = 0.9350) with controlled variation (MMD = 0.0515), despite some spatial distribution differences (Wasserstein Distance = 2.0255). As visualized in Figure~\ref{fig:instruction_analysis}(a) using UMAP dimensionality reduction and Gaussian kernel density estimation, our method maintains the core distribution while introducing beneficial diversity, demonstrating strong distributional fidelity with controlled variations. 

\paragraph{Categorical Expansion} We employ intent tagging to analyze task type distributions as introduce in section~\ref{sec:instruction_synthesis}. While seed data covers 8 primary planning categories, our synthesis expands coverage to 15 categories. Notably, this expansion maintains proportional representation of core planning tasks, preventing categorical drift.

\paragraph{Quality Preservation} Professional urban planners evaluated 100 randomly sampled instruction-response pairs from both seed and synthesized data across three dimensions (0-1 scale): planning expertise, factual correctness, and fluency. Synthesized instructions maintained comparable quality to expert-created seed data (planning expertise: 0.87 vs. 0.89; correctness: 0.85 vs. 0.88; fluency: 0.91 vs. 0.90), with no statistically significant differences (p>0.05). This confirms our approach preserves expert annotation quality while achieving substantial scale improvements.

\subsection{Preserving General Capabilities While Enhancing Domain Expertise}  \label{sec:domain_general_analysis}

\begin{figure}[t]
   \centering
   \begin{subfigure}[b]{0.48\columnwidth}
       \includegraphics[width=\textwidth]{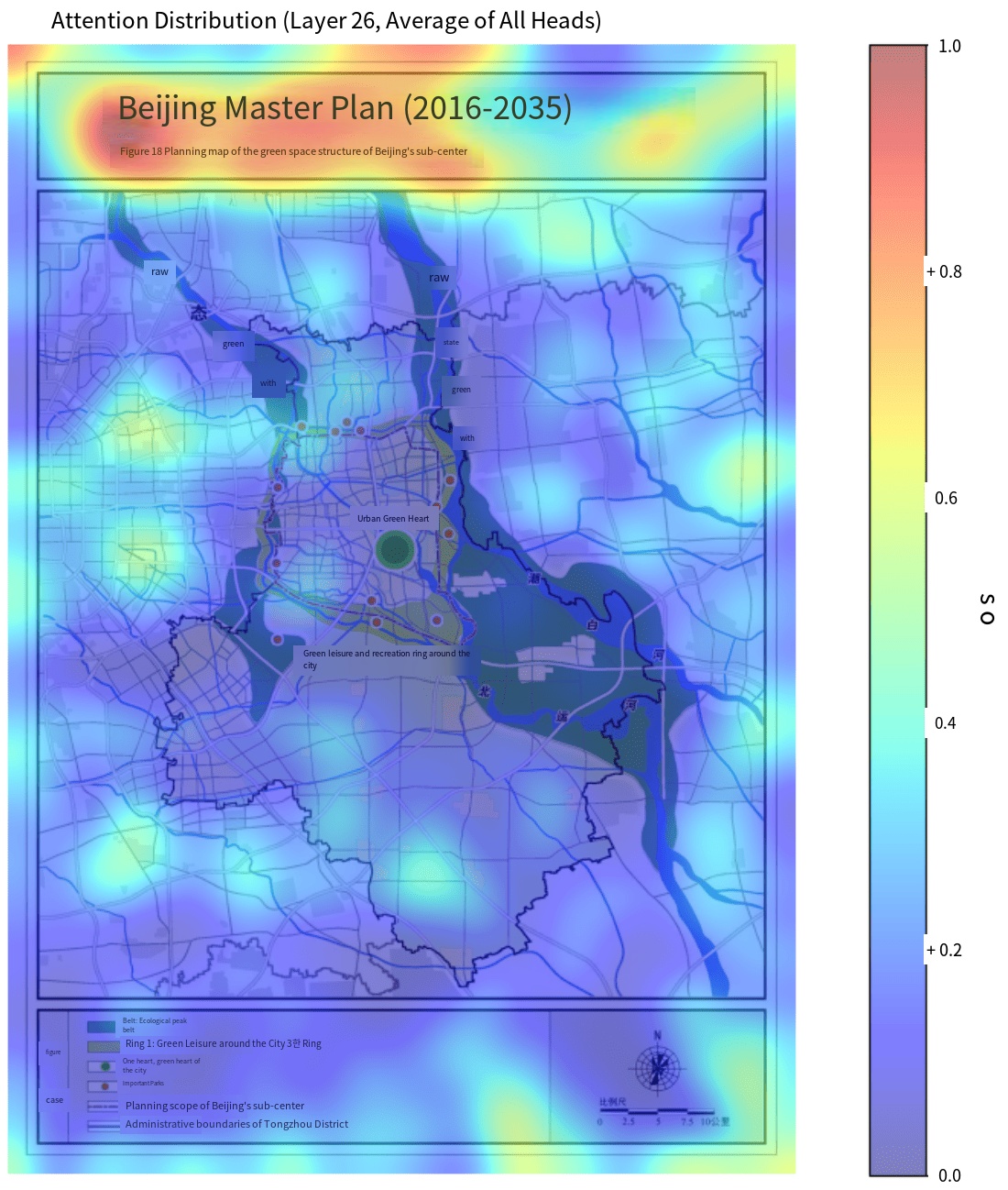}
       \caption{with configuration}
       \label{fig:attention_with_caption}
   \end{subfigure}
   \hfill
   \begin{subfigure}[b]{0.48\columnwidth}
       \includegraphics[width=\textwidth]{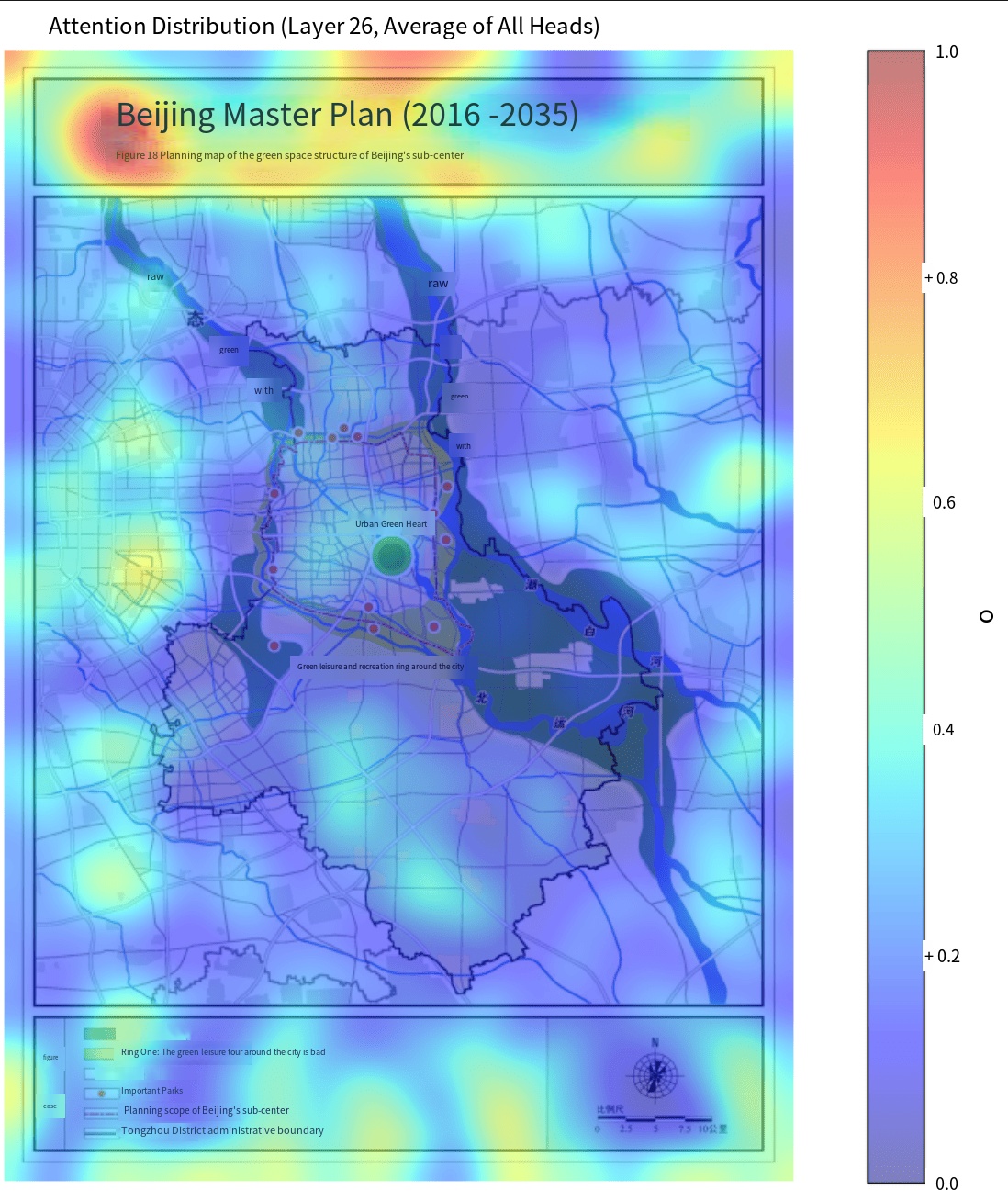}
       \caption{w/o configuration}
       \label{fig:attention_without_caption}
   \end{subfigure}
   \caption{Attention visualization comparing models with and without caption integration.}
   \label{fig:attention_comparison}
\end{figure}

We investigate how to prevent general visual capability degradation while enhancing planning specialization. We explore three key strategies: (1) Data Mixing: incorporating 5k examples from ShareGPT4-V \cite{chen2024sharegpt4v}; (2) Architecture Modifications: unfreezing the vision encoder; and (3) Caption Training: including or removing caption data. Table~\ref{tab:domain_general_analysis} presents our findings across both planning expertise (PlanBench-V) and general visual understanding (MMMU). Our analysis reveals that the optimal configuration combines mixed-domain data, a frozen vision encoder, and caption integration. As shown in Figure~\ref{fig:attention_comparison}, models trained without this configuration exhibit collapsed attention patterns, losing the interpretable focus of the base model. This attention degradation correlates with reduced performance on general visual tasks, reflecting catastrophic forgetting. Our findings demonstrate that a balanced approach with mixed training data and caption integration effectively preserves general visual capabilities while enhancing domain-specific expertise.

\begin{table}[t]
\centering
\resizebox{\columnwidth}{!}{
\begin{tabular}{clcccc}
\toprule
& \textbf{Model} & \textbf{\# Training Data} & \textbf{Planning Skill} & \textbf{General Skill} & \textbf{Avg} \\
\midrule
1 & Qwen2-VL-7B-Instruct        & - & 0.910 & \textbf{51.6} & \textbf{26.26} \\
2 & \mname                  & 11k & 1.566 & 45.8 & 23.68 \\
3 & 2 (w/ Mix Data)  & \textbf{16k} & 1.52 & 47.3 & 24.41 \\
4 & 2 (unfreeze Vision tower) & 11k & 1.54 & 44.3 & 22.92 \\
5 & 2 (w/o caption)    & 10k  & \textbf{1.59} & 45.4 & 23.50 \\
\bottomrule
\end{tabular}
}
\caption{Comparison of training configurations showing the trade-off between planning expertise and general visual understanding. }
\label{tab:domain_general_analysis}
\end{table}

\subsection{Base Model Architecture and Data Scaling Analysis}

Following the configuration in Section \ref{sec:experiments}, we analyze \mname's performance across architectures and training data scales (Table \ref{tab:combined_analysis}). Qwen2-VL-7B-Instruct achieves highest performance (1.566), while LLaVA-1.5-7B shows largest improvement (+127.5\%). Smaller models demonstrate substantial gains, suggesting effective domain specialization across architectures. For data scaling, performance improves dramatically from 100 images (-4.1\%) to 500 images (+73.6\%), then stabilizes with 1,000 images (+72.1\%). This demonstrates that our PlanAnno-V framework consistently improves performance across models of different sizes and language foundations, once a minimum data threshold is reached, highlighting the effectiveness and transferability of our approach.

\begin{table}[t]
\centering
\resizebox{\columnwidth}{!}{
\begin{tabular}{lcccc}
\toprule
\multicolumn{5}{c}{\textbf{Base Model Analysis}} \\
\midrule
\textbf{Model} & \textbf{Params} & \textbf{Original} & \textbf{Ours} & \textbf{Improv.} \\
\midrule
Qwen2-2B & 2B & 0.731 & 1.352 & +85.0\% \\
Qwen2-7B & 7B & 0.910 & \textbf{1.566} & +72.1\% \\
LLaVa-7B & 7B & 0.171 & 0.389 & \textbf{+127.5\%} \\
LLaVa-13B & 13B & 0.223 & 0.474 & +113.6\% \\
\midrule
\multicolumn{5}{c}{\textbf{Training Data Analysis}} \\
\midrule
\textbf{Data Size} & \textbf{Params} & \textbf{Original} & \textbf{Ours} & \textbf{Improv.} \\
\midrule
100 imgs & 7B & 0.910 & 0.873 & -4.1\% \\
500 imgs & 7B & 0.910 & \textbf{1.580} & \textbf{+73.6\%} \\
1,000 imgs & 7B & 0.910 & 1.566 & +72.1\% \\
\bottomrule
\end{tabular}
}
\caption{Analysis of model architectures and training data configurations.}
\label{tab:combined_analysis}
\end{table}

\section{Conclusions} \label{Conclusions}
In this paper, we introduced \mname, the first domain-specific Vision-Language Model tailored for urban planning map interpretation. Through our PlanAnno-V framework and Critical Point Thinking methodology, we efficiently addressed the challenges of data scarcity and hallucination reduction in this specialized domain. Our experiments demonstrate that \mname outperforms general-purpose VLMs by an average of 59.2\% on specialized tasks, with our lightweight 7B parameter model achieving comparable performance to models exceeding 72B parameters. This research advances AI applications in urban planning and provides a blueprint for developing specialized VLMs in other domains. \mname offers planners, policymakers, and educators a reliable tool for map analysis and decision support, enhancing both professional practice and public engagement in urban planning.

\section{Limitations}
Despite \mname's improvements, several limitations remain. While our Critical Point Thinking approach substantially reduces hallucinations, complete elimination of factual errors remains challenging, particularly for complex planning maps with ambiguous visual elements or when interpreting multiple scales simultaneously. Additionally, our approach requires a trade-off between domain specialization and general capabilities, as evidenced by performance degradation on general benchmarks. Our model's effectiveness is also constrained by training data diversity, with current implementation primarily focused on Chinese urban planning contexts. Future work should address these limitations through enhanced verification mechanisms, balanced training strategies, and expanded cross-cultural planning data.

\clearpage
\newpage

\bibliography{custom}

\clearpage

\appendix
\section{Appendix}

\subsection{Domain-Specific Data Preprocessing Details} \label{appendix:data_preprocessing}

Our data preprocessing pipeline involves several sophisticated steps to ensure high-quality planning maps for annotation and model training:

\paragraph{Map Collection and Extraction} We collected approximately 5,000 master plans and detailed planning maps from urban planning bureaus across China. These documents were primarily in PDF format, requiring specialized extraction techniques. We employed PDF parsers with custom configurations to extract high-resolution visual content while preserving spatial relationships and annotations critical to planning interpretation.

\paragraph{Quality Filtering} Initial filtering employed a multi-stage approach:
\begin{itemize}
    \item Resolution-based filtering: We established minimum resolution thresholds (1000×1000 pixels) to ensure sufficient detail for fine-grained planning elements.
    \item Information density assessment: We used computer vision techniques to quantify the information content of each map, filtering out overly sparse or dense representations.
    \item LLM-as-judge evaluation: We designed specialized prompts (detailed in Appendix \ref{appendix:filter_image_prompt}) that enabled large language models to assess information density, clarity, and planning relevance of extracted maps.
\end{itemize}


This rigorous preprocessing approach ensured our seed dataset represented authentic professional planning expertise while maintaining high visual and informational standards.

\subsection{Filter Image Prompts} \label{appendix:filter_image_prompt}

\begin{figure}[h]
\centering
\begin{tcolorbox}[, title=Image Filter Prompt(Part 1)]
PLANNING\_MAP\_PROMPT = """You are an urban planning expert. Please determine if the image below is a "complete and independent urban or territorial spatial planning map."

Please first provide a brief description of the image, then determine if it is a planning

\end{tcolorbox}
\label{fig:prompt_filter}
\end{figure}

\begin{figure}[h]
\centering
\begin{tcolorbox}[, title=Image Filter Prompt(Part 2)]
 map.

Judgment criteria:
1. It must be a complete planning map, not a part or screenshot of a planning map
2. The planning map should be the main content of the image, occupying the main area of the image
3. It should not contain many page elements unrelated to the planning map (such as large text descriptions, tables, etc.)
4. It should have typical features of planning maps:
   - Clear visual structure of spatial layout or land zoning
   - Essential planning map elements such as legend, scale, direction indicator, map title, planning unit, etc.

Please note:
- If the image is a scan of an entire document page, and the planning map is only part of the page, it should be determined as not meeting the requirements
- If the image contains multiple planning maps, it should also be determined as not meeting the requirements
- If the image content is blurry and difficult to recognize the planning content, it should be determined as not meeting the requirements

After analyzing the image, please output in the following format:

Analysis: [Provide your analysis here]
Determination: If it is a complete and independent planning map, please output: \boxed{1}
If it is not a complete and independent planning map, please output: \boxed{0}
"""

\end{tcolorbox}
\end{figure}


\subsection{Data Analysis} \label{appendix:data_analysis}
\begin{figure}[H]
  \centering
  \includegraphics[width=0.45\textwidth]{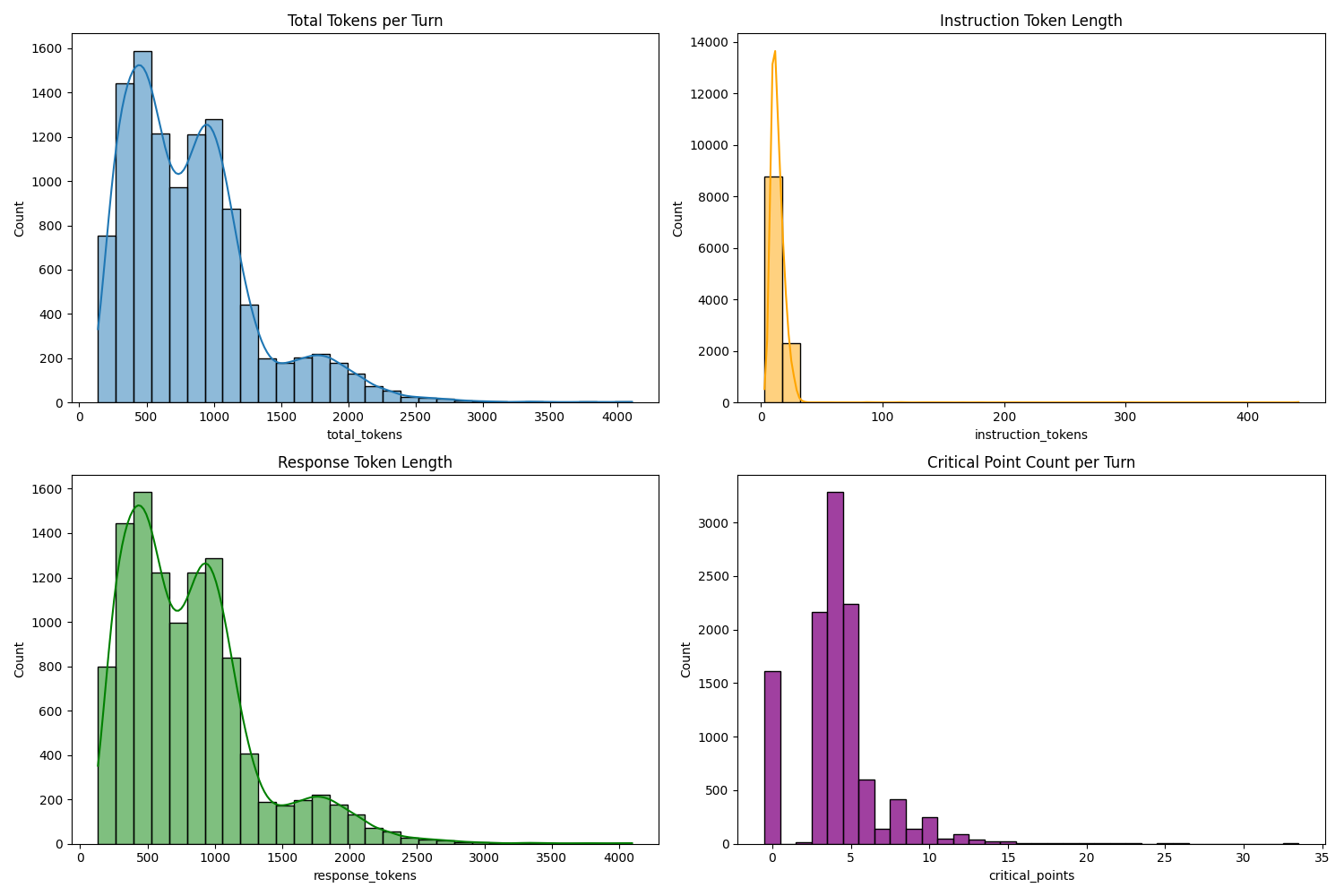}
  \caption{Token Distribution and Cirtical Point Distribution Analysis}
  \label{fig:token_distribution_analysis}
\end{figure}

We conduct a comprehensive analysis of the dialogue dataset from two perspectives:

First, we analyze the token distribution across each round of dialogue, including the number of tokens in the instruction, response, and their total. This provides insights into the input-output complexity of the dataset. Additionally, we examine the distribution of critical points per round to understand the density of semantic shifts or decision points in the dialogues. The results are shown in Figure \ref{fig:token_distribution_analysis}.

\begin{figure}[H]
  \centering
  \includegraphics[width=0.5\textwidth]{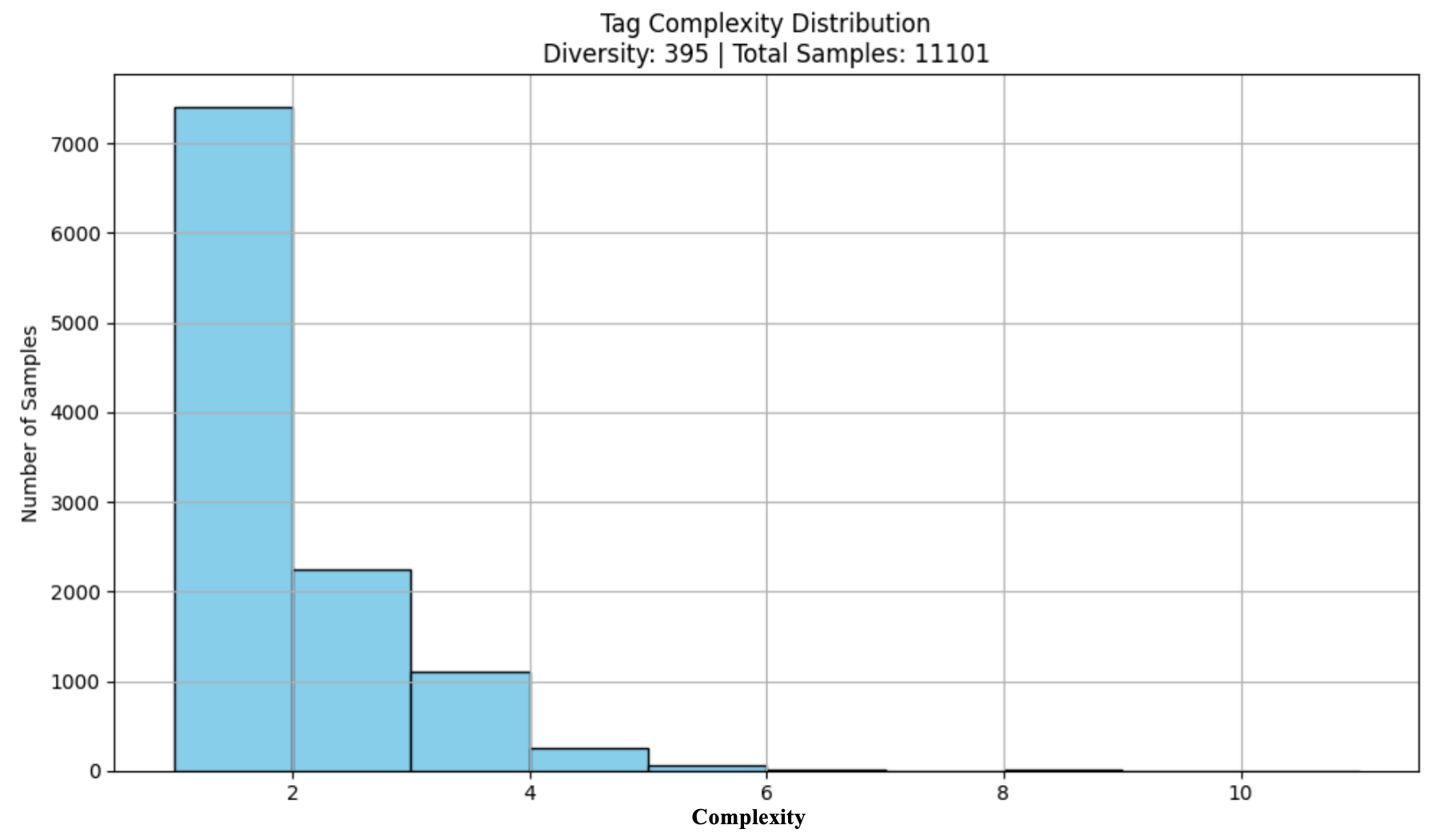}
  \caption{Instagger Analysis}
  \label{fig:instagger_analysis}
\end{figure}

Second, to assess the semantic diversity and complexity of the instructions, we employ the Instagger model to map each instruction into a predefined tag space. This allows us to analyze the diversity of task types and compute the number of tags associated with each instruction to estimate its semantic complexity. The corresponding analysis is illustrated in Figure \ref{fig:instagger_analysis}.

\subsection{Comparison of PlanGPT-VL and Qwen}

\begin{figure*}[htbp]
\centering
\begin{tcolorbox}[, title=Example 1]
\textbf{Question}Please describe the national key wetlands in the picture.(image see Fig~\ref{fig:appendix_example1_fig})\\

\textbf{PlanGPT-VL}: According to the legend, the National Key Wetland is represented by a light gray shadow. It is located on the southern side of the planned expressway, near the river (within the river embankment) area. The wetland is situated close to the county town and the food industry park, but maintains a certain distance from them, thereby avoiding direct disturbance. The primary functions of the National Key Wetland include protecting biodiversity, regulating the climate, conserving water resources, and purifying water quality. At the same time, it serves as an important ecological buffer zone for the city, preventing excessive interference with natural ecosystems caused by urban expansion. In urban planning, the establishment of the National Key Wetland reflects the coordination between ecological protection and urban development, contributing to the harmonious coexistence between the city and nature.\\
\textbf{Qwen2-7B-Instruct}: According to the information in the picture, the national key wetlands are located in the lower right corner of the map and are represented by light gray areas.\\
\textbf{Qwen2-72B-Instruct-AWQ}:In this map, the national key wetland is marked as a small area near the river. This area is labeled "National Key Wetland" and outlined with a dashed frame.
\end{tcolorbox}
\label{fig:appendix_example1_box}
\end{figure*}

\begin{figure*}[htbp]
\centering
\includegraphics[width=0.9\textwidth]{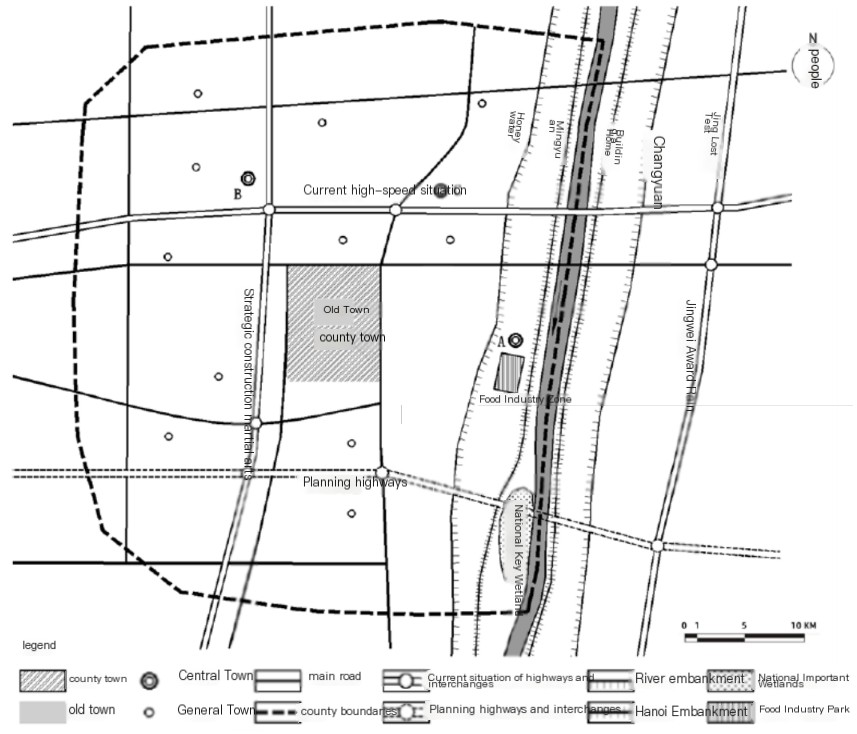}
\caption{Image of Example 1}
\label{fig:appendix_example1_fig}
\end{figure*}

\begin{figure*}[htbp]
\centering
\begin{tcolorbox}[, title=Example 2]
\begin{CJK*}{UTF8}{gbsn}
\textbf{Question}: 世界遗产在地图上是如何标识的？北京市有哪些世界遗产被列入了这张地图？(image see Fig~\ref{fig:appendix_example2_fig})

\textbf{PlanGPT-VL}: 世界遗产在地图上以红色圆点表示，用于标识其位置。北京市的世界遗产包括：故宫、天坛、颐和园、周口店北京人遗址、长城、大运河、西山永定河文化带（如妙峰山、潭柘寺等）、大运河文化带（如通州大运河、元大都遗址等）。这些世界遗产被纳入北京市市域历史文化名城保护结构规划中，体现了北京市对世界遗产的高度重视和保护。

\textbf{Qwen2-7B-Instruct}: 世界遗产在地图上用紫色圆圈标识。北京市被列入这张地图的世界遗产有：故宫、天坛、颐和园、周口店北京人遗址、长城、大运河。

\textbf{Qwen2-72B-Instruct-AWQ}: 世界遗产在地图上是用红色圆圈标识的。北京市被列入这张地图的世界遗产包括：长城、故宫、天坛、颐和园、十三陵、周口店北京人遗址、明十三陵、清东陵、清西陵、圆明园、北海公园、景山公园、中山公园、天安门广场、北京动物园、北京植物园、北京天文馆、北京自然博物馆、北京科技馆、北京艺术博物馆、北京历史博物馆、北京军事博物馆、北京革命博物馆、北京民俗博物馆、北京工艺美术博物馆、北京古建筑博物馆、北京古生物博物馆、北京古陶瓷博物馆、北京古钱币博物馆、北京古玉器博物馆、北京古家具博物馆、北京古钟表博物馆、北京古乐器博物馆、北京古书画博物馆、北京古籍博物馆、北京古籍修复博物馆、北京古籍装裱博物馆、北京古籍装订博物馆、北京古籍装订工具博物馆、北京古籍装订材料博物馆、北京古籍装订技术博物馆、北京古籍装订历史博物馆、北京古籍装订艺术博物馆、北京古籍装订工艺博物馆...
\end{CJK*}
\end{tcolorbox}
\label{fig:appendix_example2_box}
\end{figure*}

\begin{figure*}[htbp]
\centering
\includegraphics[width=0.55\textwidth]{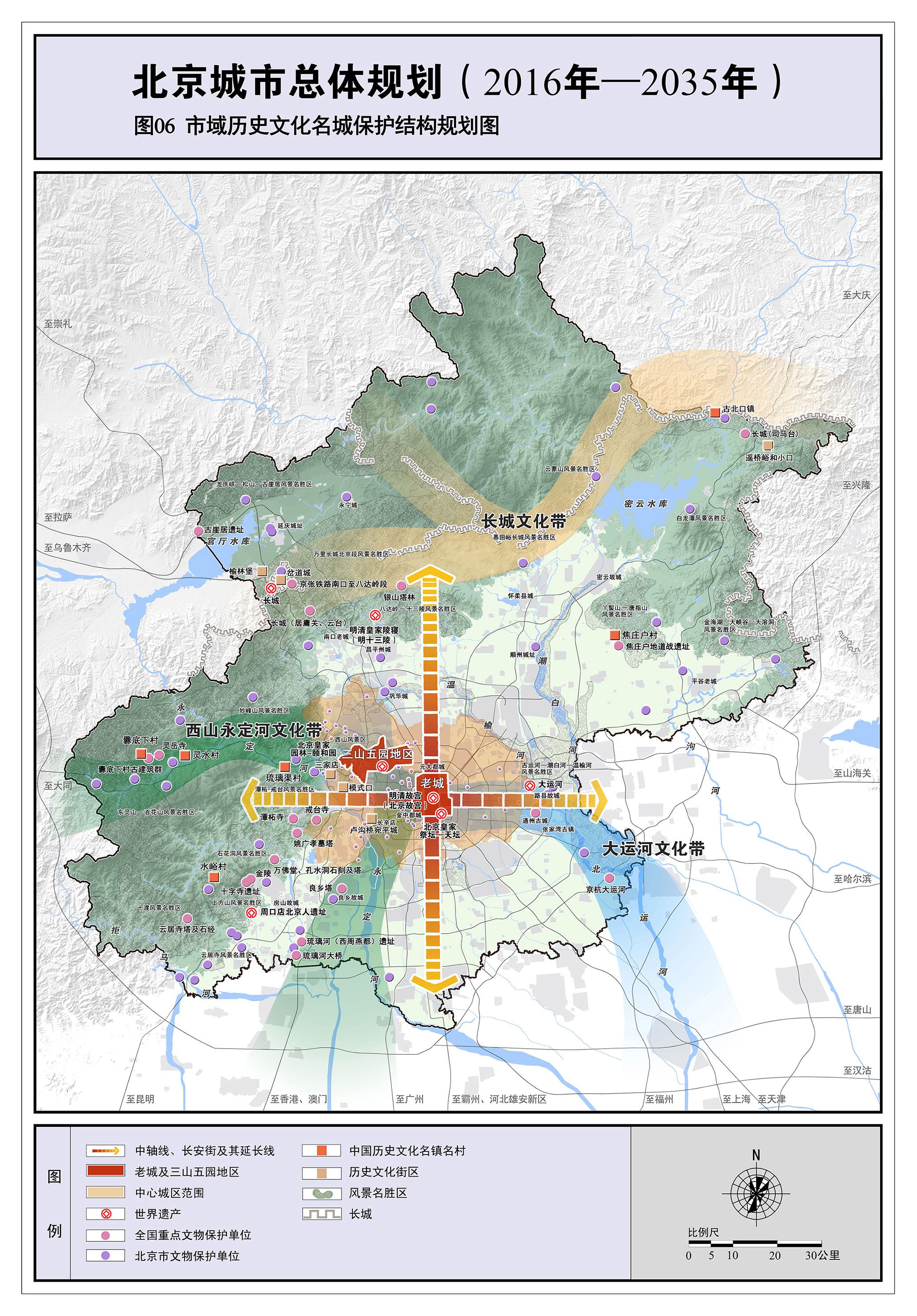}
\caption{Image of Example 2}
\label{fig:appendix_example2_fig}
\end{figure*}

\begin{figure*}[htbp]
\centering
\begin{tcolorbox}[, title=Example 3]
\textbf{Question} How is the transportation connection between the "Beijing New Airport" and the surrounding areas in the image? (image see Fig~\ref{fig:appendix_example3_fig})\\

\textbf{PlanGPT-VL}: Translation (without changing the format):

---

According to the information in the image, the transportation connection between Beijing New Airport and the surrounding areas is as follows:

1. **Location of Beijing New Airport**:
   Beijing New Airport is located in the lower right corner of the image, near Daxing District. It serves as an important transportation hub in the image.

2. **Surrounding Transportation Lines**:
   * **Operational Lines**: The image shows that there are operational lines near Beijing New Airport, including Line 8, Line S1, etc. These lines directly connect Beijing New Airport with the central urban area and surrounding regions.
   * **Planned Lines**: The image also indicates planned lines (including those under construction), such as the southern and northern extensions of Line S1. These lines will further improve the transportation network of Beijing New Airport.

3. **Transfer Stations and Regular Stations**:
   * **Transfer Stations**: There are several transfer stations around Beijing New Airport, such as Daxing Airport Station and Daxing Airport North Station. These stations enable efficient transfers between rail transit and surface transportation.
   * **Regular Stations**: There are also regular stations around Beijing New Airport, such as Daxing Airport Station and Daxing Airport North Station, providing passengers with convenient transfer and travel services.

4. **Transportation Accessibility**:
   * **Distance**: According to the scale in the image, Beijing New Airport is relatively close to the central urban area and other regions, indicating good transportation accessibility.
   * **Transfer Convenience**: The distribution of transfer and regular stations around Beijing New Airport is reasonable, enabling seamless integration of rail and surface transportation and enhancing passengers' transfer experience.

5. **Conclusion**:
   * The transportation connection between Beijing New Airport and the surrounding areas is relatively strong. The combination of operational and planned lines enables efficient linkage with the central urban area and surrounding regions.
   * The layout of transfer and regular stations is well-designed to meet passengers' transfer needs and improve overall transportation accessibility.

In summary, the transportation connection between Beijing New Airport and the surrounding areas is well-developed. With the coverage of the rail transit network and the layout of transfer stations, it enables convenient transportation accessibility, supporting the efficient operation of Beijing New Airport and coordinated regional development.\\

\textbf{Qwen2-7B-Instruct}: The transportation connection between "Beijing New Airport" and the surrounding areas in the image is mainly achieved through rail transit. As shown in the image, there are multiple rail transit lines around Beijing New Airport, including Line 8, Line 19, and the New Airport Line. These lines have stations near the new airport, providing convenient access for passengers. In addition, the New Airport Line connects to the central area of Beijing, further strengthening the transportation link between the new airport and the urban center.\\

\textbf{Qwen2-72B-Instruct-AWQ}: The image shows that the transportation connection between "Beijing New Airport" and the surrounding areas is very close. Multiple rail transit lines link the new airport with the city center of Beijing and surrounding regions. Specifically, there are several blue lines near the new airport, which represent planned lines, including those under construction. These lines are connected to other routes through transfer stations, forming a convenient transportation network. Through these lines, passengers can quickly reach the city center of Beijing and various surrounding areas from the new airport.
\end{tcolorbox}
\label{fig:appendix_example3_box}
\end{figure*}

\begin{figure*}[htbp]
\centering
\includegraphics[width=0.9\textwidth]{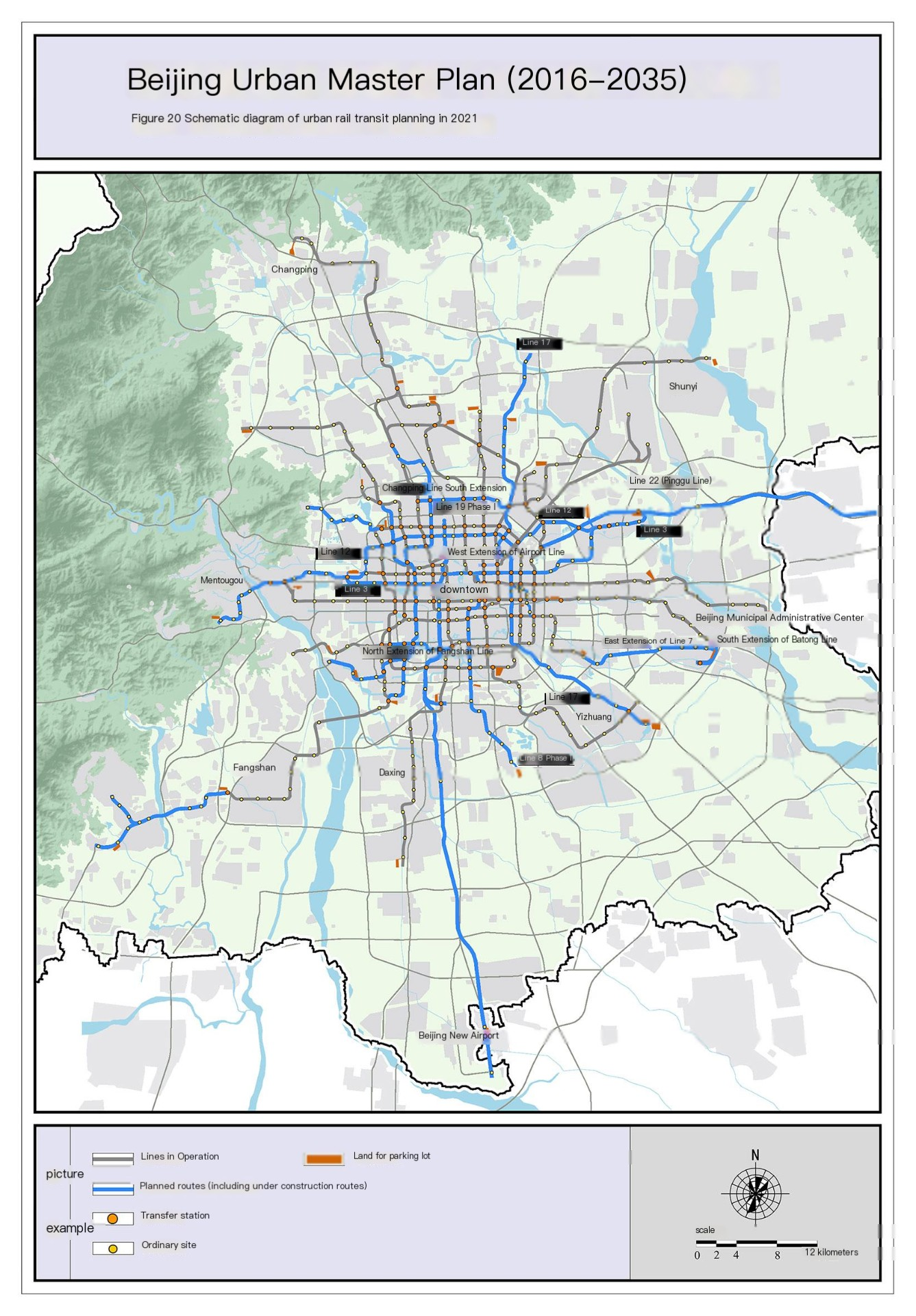}
\caption{Image of Example 3}
\label{fig:appendix_example3_fig}
\end{figure*}

\subsection{Results of General Benchmark}


\subsection{Coompare Attention Score Map}

\begin{figure*}[htbp]
\centering
\begin{subfigure}[t]{0.48\textwidth}
    \centering
    \includegraphics[width=\textwidth]{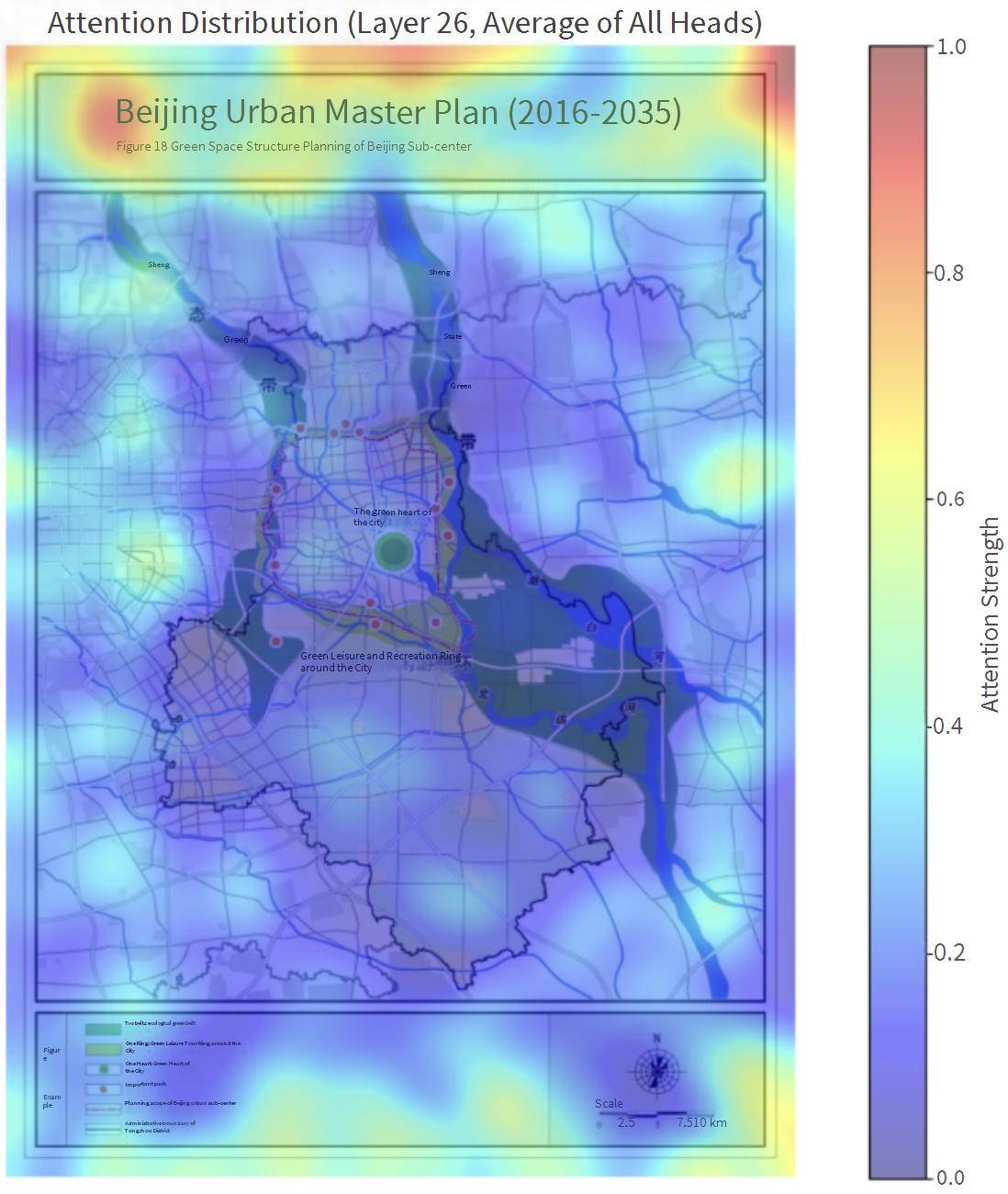}
    \caption{Base Model}
\end{subfigure}
\hfill
\begin{subfigure}[t]{0.48\textwidth}
    \centering
    \includegraphics[width=\textwidth]{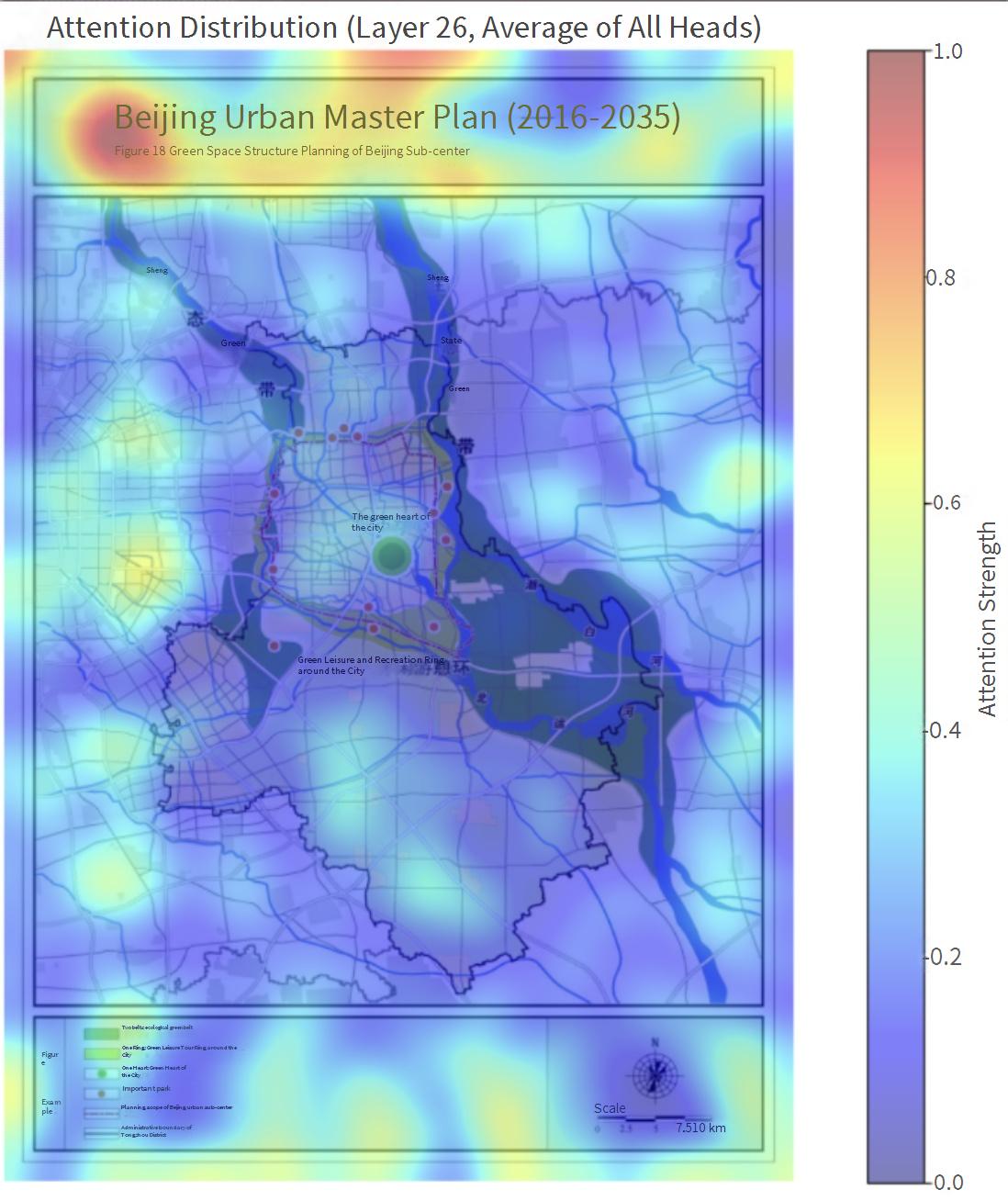}
    \caption{Mix Model}
\end{subfigure}
\caption{Attention scores for question: Where is the green heart of the city?}
\label{fig:green_heart}
\end{figure*}

\begin{figure*}[htbp]
\centering
\begin{subfigure}[t]{0.48\textwidth}
    \centering
    \includegraphics[width=\textwidth]{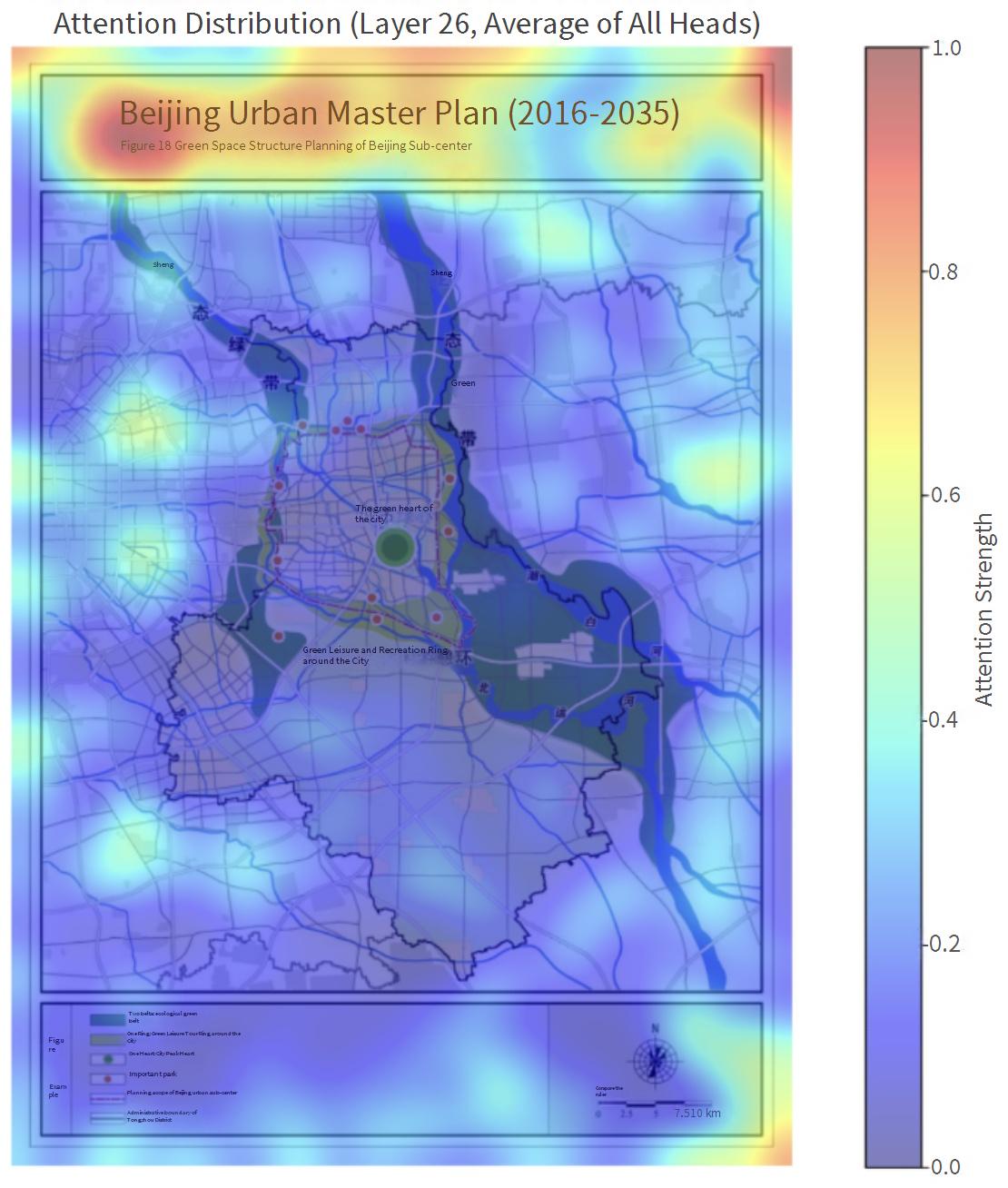}
    \caption{Base Model}
\end{subfigure}
\hfill
\begin{subfigure}[t]{0.48\textwidth}
    \centering
    \includegraphics[width=\textwidth]{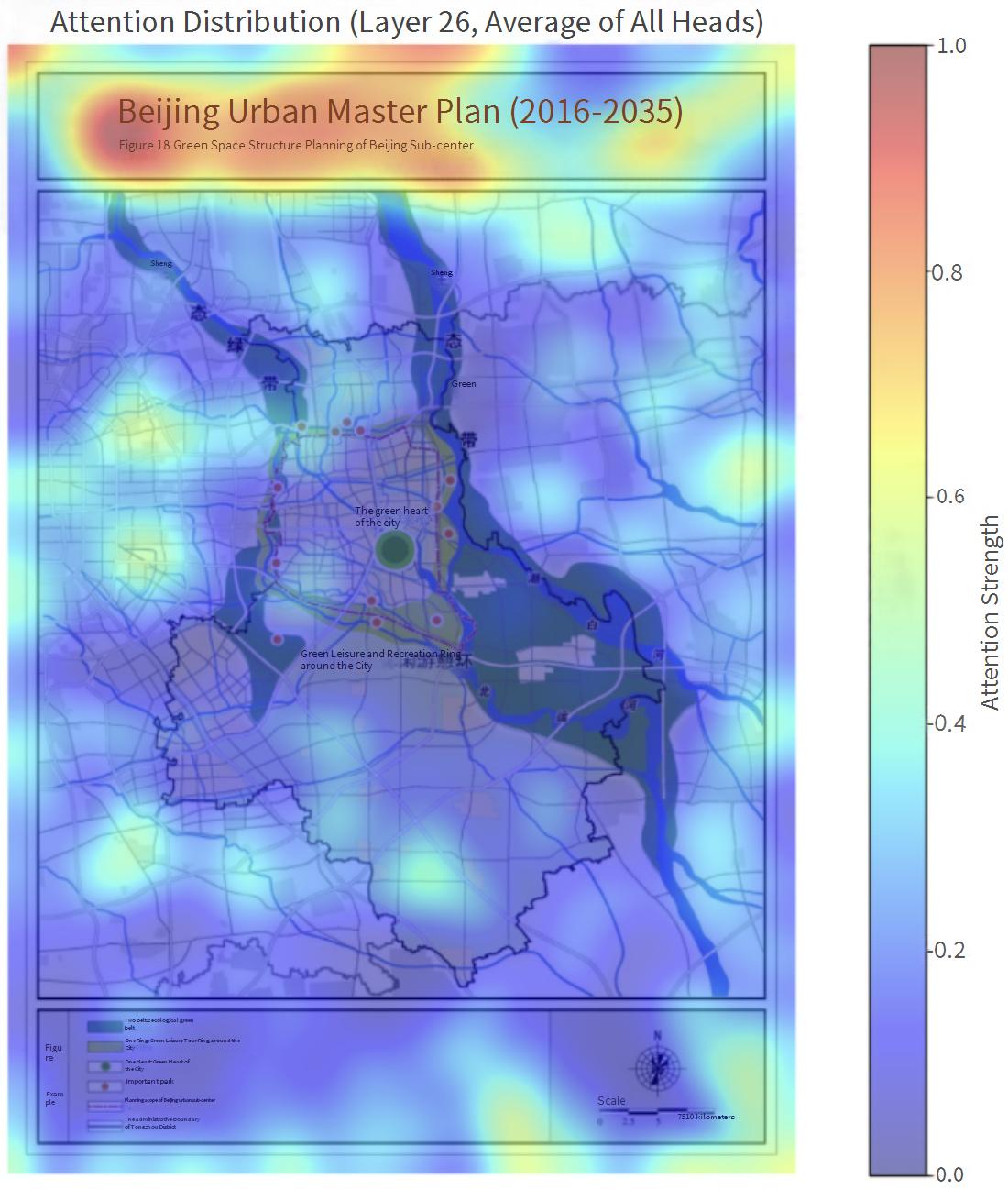}
    \caption{Mix Model}
\end{subfigure}
\caption{Attention scores for question: Please describe this image}
\label{fig:describe_image}
\end{figure*}

\begin{figure*}[htbp]
\centering
\begin{subfigure}[t]{0.48\textwidth}
    \centering
    \includegraphics[width=\textwidth]{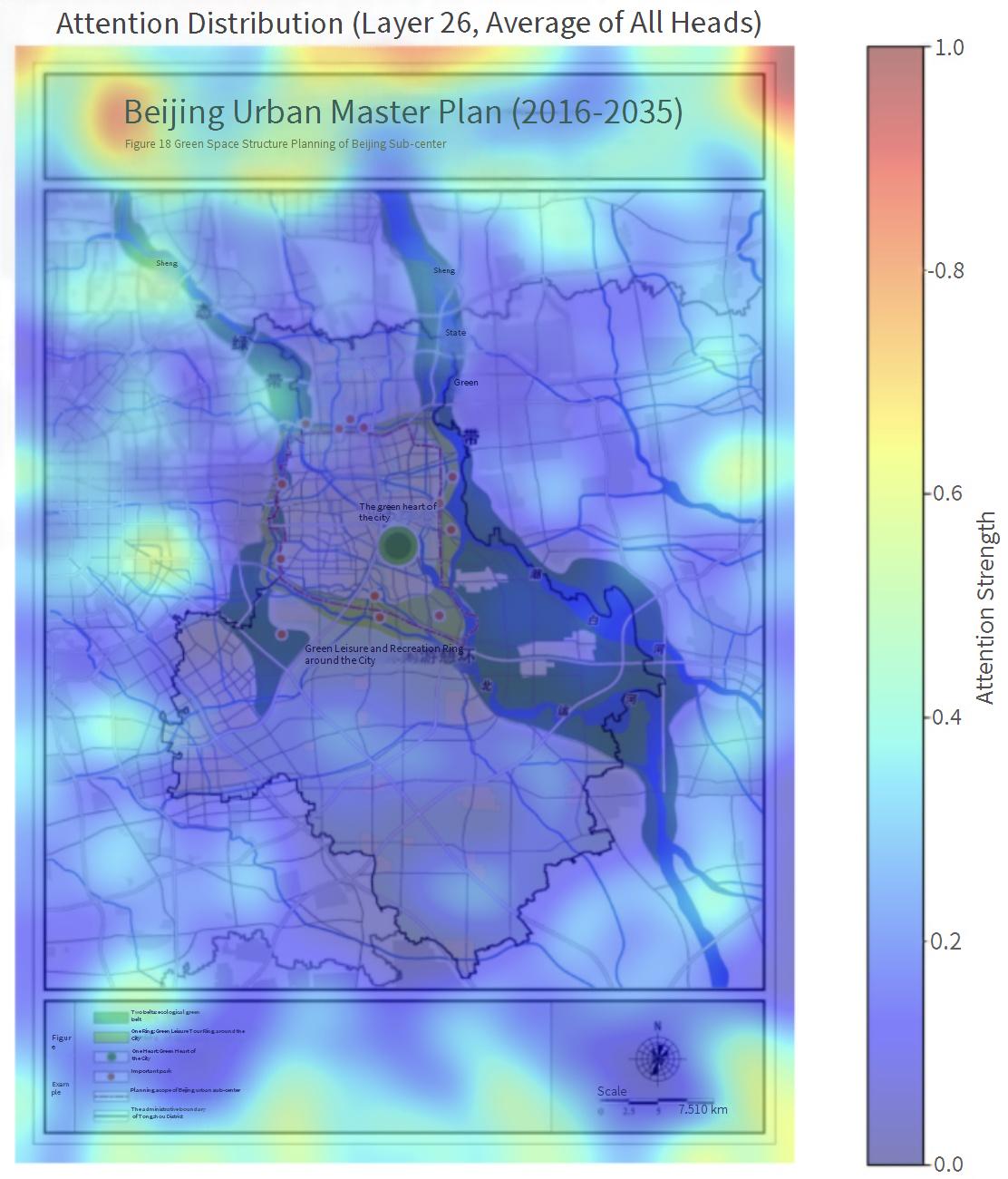}
    \caption{Base Model}
\end{subfigure}
\hfill
\begin{subfigure}[t]{0.48\textwidth}
    \centering
    \includegraphics[width=\textwidth]{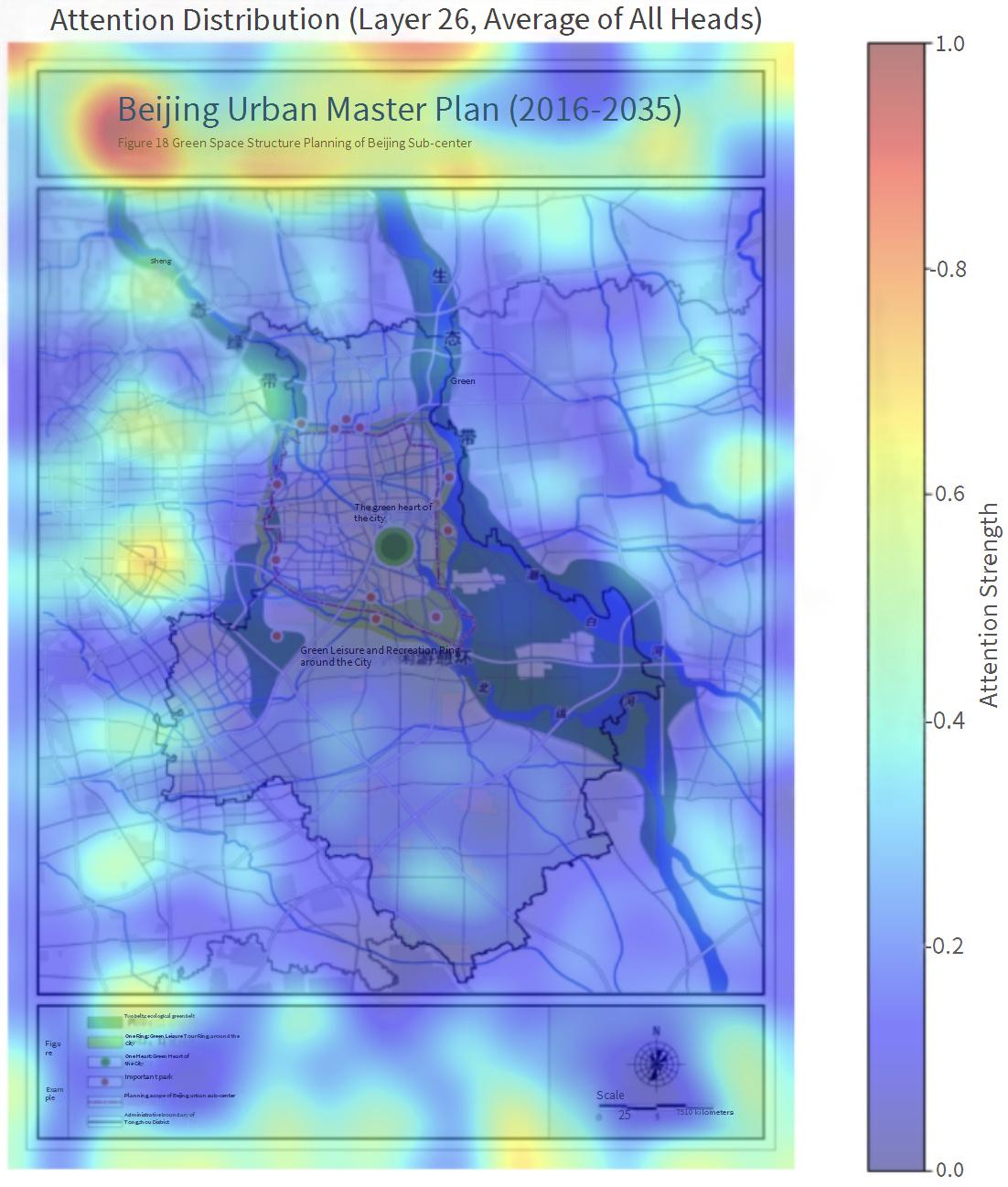}
    \caption{Mix Model}
\end{subfigure}
\caption{Attention scores for question: Where is the ecological green belt}
\label{fig:eco_belt}
\end{figure*}

\subsection{Evaluation Prompt}
\label{appendix:planvlm_eval}

\begin{figure*}[htbp]
\centering
\begin{tcolorbox}[, title=Prompt of Evaluation]
evaluation\_prompt = f"""
Please evaluate the following answer based on the question, the list of scoring points, and the image content.

Question: $\{question\} $

Scoring Points List: $\{critical\_points\_text\}$

Answer to be Evaluated: $\{summary\}$

Scoring Criteria:
Please check whether the model's answer covers each scoring point:
- For each scoring point, if the model's answer includes relevant content, give 1 point.
- If the scoring point is not covered or is incorrectly described, give 0 points.
- The scoring points are mutually exclusive, and each point can earn a maximum of 1 point.

Please score according to the following format:
1. Scoring Point 1: [0/1] – Briefly explain whether this point is covered and provide reasoning
2. Scoring Point 2: [0/1] – Briefly explain whether this point is covered and provide reasoning
...

Final Score: X/Y (X is the total score, Y is the total number of scoring points)
"""

\end{tcolorbox}
\label{fig:prompt_evaluation}
\end{figure*}

\subsection{Evaluation Example}
\label{appendix:planvlm_eval_example}

\begin{figure*}[htbp]
\centering
\begin{tcolorbox}[, title=Evaluation Example 1 (Part 1) ]
\textbf{Question}: \\
How does Jiangsu Province promote urban-rural integration in territorial spatial planning?

\textbf{Answer}: \\
In its territorial spatial planning, Jiangsu Province promotes urban-rural integration by scientifically delineating the urban development boundary. The plan explicitly states that the expansion multiplier of the urban development boundary is controlled within 1.3 times the scale of urban construction land in 2020, ensuring orderly urban development. At the same time, by strengthening infrastructure construction, the province promotes integrated urban-rural development and narrows the urban-rural gap. Additionally, Jiangsu optimizes the functional layout of towns to improve their overall carrying capacity, providing a higher quality living environment for urban and rural residents.

\textbf{Critical Points}:
\begin{itemize}
  \item [1] Scientifically delineate urban development boundaries, controlling the expansion multiplier within 1.3 times.
  \item [2] Ensure orderly urban development.
  \item [3] Strengthen infrastructure construction to promote urban-rural integration.
  \item [4] Narrow the urban-rural gap.
  \item [5] Optimize town functional layout and improve overall carrying capacity.
  \item [6] Provide higher-quality living environments for urban and rural residents.
\end{itemize}

\textbf{Model Response}: \\
Jiangsu Province promotes urban-rural integration in territorial spatial planning mainly through the following measures:

1. \textbf{Comprehensive Planning}: Develop integrated urban-rural spatial development plans to ensure coordinated development in functions, infrastructure, and public services between cities and rural areas.

2. \textbf{Infrastructure Construction}: Increase investment in rural infrastructure, improving transportation, water supply, electricity, and internet access to enhance rural living standards.

3. \textbf{Land Use Policy}: Optimize land use structure through scientific zoning and intensive utilization to rationally allocate agricultural and construction land.

4. \textbf{Industrial Transformation and Upgrading}: Encourage industrial coordination between urban and rural areas, support the development of characteristic rural industries, and boost rural economic vitality.

5. \textbf{Ecological Protection}: Strengthen environmental protection and promote green development to ensure harmony between development and nature.

6. \textbf{Policy Support}: Implement policies favorable to urban-rural integration, including financial support and tax incentives to stimulate investment and innovation.

7. \textbf{Social Services}: Improve rural education, healthcare, and cultural services to raise living standards and attract talent back to rural areas.

In summary, Jiangsu’s approach to urban-rural integration aims to achieve comprehensive and harmonious development through coordinated planning and multifaceted policy support.

\end{tcolorbox}
\label{fig:evaluation_example}
\end{figure*}

\begin{figure*}[htbp]
\centering
\begin{tcolorbox}[, title=Evaluation Example 1 (Part 2) ]
\textbf{Score Text}: \\
1. Point 1: [1] – The model mentions “comprehensive planning” and “integrated urban-rural spatial development plans,” which align with the requirement to scientifically delineate urban development boundaries.\\
2. Point 2: [1] – The response discusses coordinated development in functions, infrastructure, and public services, which corresponds to ensuring orderly urban development.\\
3. Point 3: [1] – The response clearly mentions increasing investment in rural infrastructure, which matches the point of strengthening infrastructure construction.\\
4. Point 4: [1] – Improving education, healthcare, and cultural services in rural areas helps narrow the urban-rural gap.\\
5. Point 5: [1] – Scientific zoning and rational land allocation reflect optimizing town functions and improving carrying capacity.\\
6. Point 6: [0] – Although the response mentions raising living standards, it does not explicitly emphasize providing a higher-quality living environment, and thus does not fully meet this point.\\

Final Score: 5/6
\end{tcolorbox}
\end{figure*}

\begin{figure*}[htbp]
\centering
\includegraphics[width=0.9\textwidth]{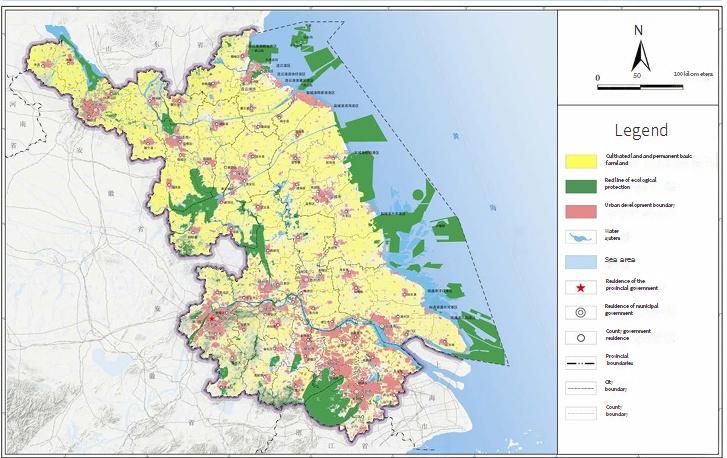}
\caption{Image of Evaluation Example 1}
\label{fig:appendix_eval_example1}
\end{figure*}

\begin{figure*}[htbp]
\centering
\begin{tcolorbox}[, title=Evaluation Example 2]
\textbf{Question}: \\
Which category does this map belong to: master plan, regulatory detailed plan, construction detailed plan, or special plan?

\textbf{Answer}: \\
Construction detailed plan.

\textbf{Critical Points}:
\begin{itemize}
  \item [1] Correctly identify the map type as a "construction detailed plan."
  \item [2] Recognize the features and purpose of a construction detailed plan.
\end{itemize}

\textbf{Model Response}: \\
This map belongs to a regulatory detailed plan. Regulatory detailed plans are typically used to show specific details such as land use, road layout, and urban infrastructure.

\textbf{Score Text}: \\
1. Point 1: [0] – The model incorrectly identifies the map as a "regulatory detailed plan" instead of the correct type "construction detailed plan," thus failing to meet this point.\\
2. Point 2: [0] – The model fails to recognize the features and intended use of a construction detailed plan, so this point is not fulfilled.\\

Final Score: 0/2
\end{tcolorbox}
\label{fig:evaluation_example2}
\end{figure*}

\begin{figure*}[htbp]
\centering
\includegraphics[width=0.8\textwidth]{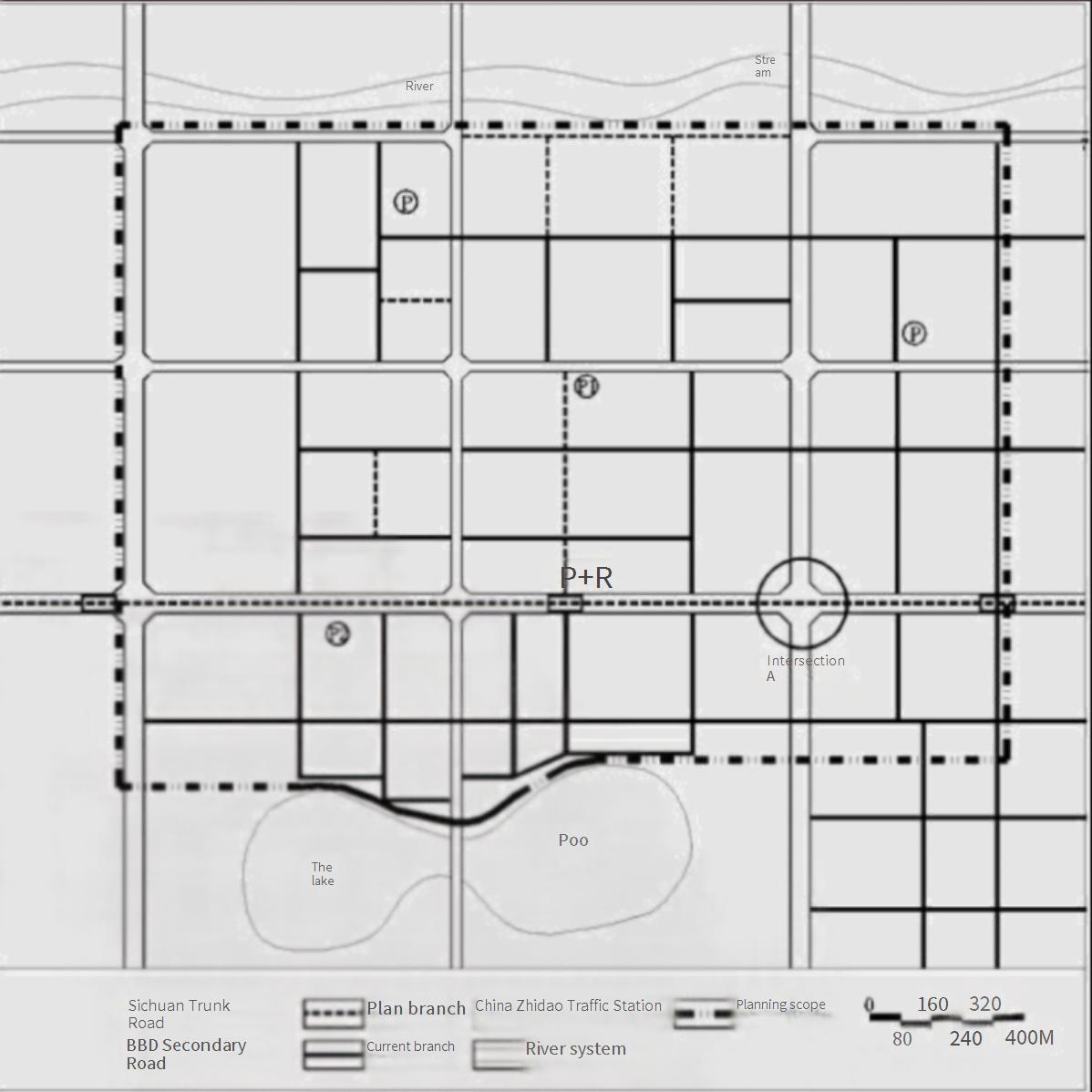}
\caption{Image of Evaluation Example 2}
\label{fig:appendix_eval_example2}
\end{figure*}

\end{document}